\documentclass[12pt]{article}
\usepackage{amsmath}
\usepackage{times}
\usepackage{graphicx}
\usepackage{color}
\usepackage{multirow}
\usepackage{amssymb}
\usepackage{booktabs}
\usepackage{url}

\usepackage{rotating}
\usepackage{latexsym}
\usepackage{subfigure} % Written by Steven Douglas Cochran
\usepackage{amsthm}
\newtheorem{definition}{Definition}

\newcommand{\captionfonts}{\normalsize}

\makeatletter
\long\def\@makecaption#1#2{%
  \vskip\abovecaptionskip
  \sbox\@tempboxa{{\captionfonts #1: #2}}%
  \ifdim \wd\@tempboxa >\hsize
    {\captionfonts #1: #2\par}
  \else
    \hbox to\hsize{\hfil\box\@tempboxa\hfil}%
  \fi
  \vskip\belowcaptionskip}
\makeatother
\begin{document}
\hspace{13.9cm}1

\ \vspace{20mm}\\

%{\LARGE Statistics of visual responses by DNN cells to IT cells in primate}

{\LARGE Face representation by deep learning: a linear encoding in a parameter space?}

\ \\
{\bf  Qiulei Dong$^{\displaystyle 1, \displaystyle 2, \displaystyle 3}$, Jiayin Sun$^{\displaystyle 1, \displaystyle 2}$, Zhanyi Hu$^{\displaystyle 1, \displaystyle 2, \displaystyle 3, *}$}\\
{$^{\displaystyle 1}$National Laboratory of Pattern Recognition, Institute of Automation, Chinese Academy of Sciences, Beijing
100190, China.}\\
{$^{\displaystyle 2}$University of Chinese Academy of Sciences, Beijing 100049, China.}\\
{$^{\displaystyle 3}$CAS Center for Excellence in Brain Science and Intelligence Technology, Beijing 100190, China.}\\
{$^*$ Corresponding author: huzy@nlpr.ia.ac.cn}\\
%

%%\ \\[-2mm]
%{\bf Keywords:} Inferotemporal Cortex, Single-Neuron Selectivity, Population Sparseness, Intrinsic Dimensionality, Deep Neural Networks (DNNs)

\thispagestyle{empty}
\markboth{}{NC instructions}
\ \vspace{-0mm}\\

%
%Abstract
%Abstract
\begin{abstract}
Recently, Convolutional Neural Networks (CNNs) have achieved tremendous performances on face recognition, and one popular perspective regarding CNNs' success is that CNNs could  learn discriminative  face representations from face images with complex image feature encoding. However, it is still unclear what is the intrinsic  mechanism of face representation in CNNs. 
In this work, we investigate this problem by formulating face images as points in a shape-appearance parameter space, and our results demonstrate that: (i) 
The encoding and decoding of the neuron responses (representations) to face images in CNNs could be achieved under a linear model in the parameter space, in agreement with the recent discovery in primate IT face neurons, but different from the aforementioned perspective on CNNs' face representation with complex image feature encoding; 
(ii) The  linear model for face encoding and decoding in the parameter space could achieve close  or even better performances on face recognition and verification than state-of-the-art CNNs, which might provide new lights on the design strategies for face recognition systems;
(iii) The neuron responses to face images in CNNs could not be adequately modelled  by the axis model, a model recently proposed on face modelling in primate IT cortex.
All these results might shed some lights on the often complained blackbox nature behind CNNs' tremendous performances on face recognition.

\end{abstract}

\section{Introduction}

Human face representation, aiming to represent the identity of human face, 
is an important and challenging topic in both the fields of computer vision and neuroscience, and  has attracted more and more attention in recent years. 

In the neuroscience field, 
visual object representation, including face representation,  is generally believed to happen in primate inferotemporal (IT) cortex, and   the population responses of IT neurons to an object image stimulus is  considered as the representation of this object  \cite{Freiwald2010,Grimaldi2016,Lehky2014,Majaj,Khaligh:PLOS,
Yamins2014Performance,Yamins2016,Chang2017,DongWH18}. 
%As for human faces, both the existing functional experiments \cite{Freiwald2010} and anatomical data \cite{Grimaldi2016} showed that the anterior medial (AM) patch in  macaque IT cortex
%is the final stage of face processing.
In the early years,  many traditional works on face representation assumed an exemplar-based mechanism for representing face identity in primate IT cortex: face identification was mediated by units tuned to a set of exemplar faces.
Such an exemplar-based representation mechanism is supported by the results in  \cite{Freiwald2010} that  some neurons in  the anterior medial face patch  are view-independent, which respond to faces of only a few specific individuals regardless of view orientations. 
Recently, different from the results in \cite{Freiwald2010},
Chang and Tsao \cite{Chang2017} found that 
by formulating face images as points in a multi-dimensional linear parameter space, face images could be linearly encoded in macaque IT cortex, and they could also be linearly decoded from IT neuron responses, and a new face representation model, called ``the axis model'', was proposed.
Their experimental results demonstrated that the proposed axis model could achieve satisfactory encoding and decoding performances of IT neuron responses.

In the computer vision field,  the performances of face recognition systems depend heavily on face representation, which is naturally coupled with many adverse factors, such as pose variation, illumination change, expression, occlusion and so on. Face representation could 
either be manually designed or automatically learnt from
face image datasets. In the early days, the  face representations  were mainly constructed with manually designed features, such as Local Binary Patterns \cite{OjalaPAMI}, Histogram
of Oriented Gradients \cite{Dalal2005Histograms}, etc. 
In recent years,
Convolutional Neural Networks (CNNs),
which are generally believed to be able to learn complex and effective representations from  image stimuli, have achieved tremendous successes on object categorization and face recognition \cite{zhu2013deep,taigman2014deepface,sun2014deep,sun2014deep1,taigman2014web,Parkhi15,
jung2015rotating,zhang2016pursuing,Zhang2017Two,Zhang2018deep,Liu_2017_CVPR,wu2018light,deng2018arcface}.
For example, DeepFace \cite{taigman2014deepface}  trained a deep CNN to classify faces using a dataset of $400,000$  examplar images.
DeepID \cite{sun2014deep} employed a CNN to learn face representations for identifying $10,000$
different faces.
In  \cite{sun2014deep1}, a new CNN was introduced to learn face representations, which was trained with both face identification and verification signals. 
Hayat et al. \cite{Hayat2017Joint} proposed a data-driven method
to jointly learn registration with face representation in a CNN.
Liu et al. \cite{Liu2017SphereFace} proposed a deep hypersphere embedding approach for face recognition, where the
angular Softmax loss for CNNs was introduced to learn discriminative face
representations (called SphereFace) with angular margin.
Zhang et al. \cite{Zhang2018deep} proposed a disentangling siamese network, which could automatically disentangle the face features into the identity representations, as well as the identity-orthogonal factors including poses and illuminations.
Wu et al. \cite{wu2018light} proposed a light CNN framework to learn a compact face representation on the large-scale face data with noisy labels.
Deng et al. \cite{deng2018arcface} proposed a geometrically interpretable loss function, called ArcFace, which is integrated with different CNN models (e.g. ResNet \cite{He7780459}) for face recognition and verification.

%\subsection{Motivation and contribution}
Why do CNNs perform so well on face recognition? One popular perspective  is that CNNs could  learn effective and discriminative face representations with complex image feature  encoding,  because of the repeatedly used  nonlinear operators such as ReLU (Rectified Linear Unit) and max pooling in CNNs. However,  \textbf{what is the intrinsic mechanism of face representation in CNNs?} It seems this is still largely an open question.
In addition, CNNs' successes in generic object categorization and recognition are often attributed by many researchers  to their inherent hierarchical architectures, similar to
the primate visual ventral pathway.
It is also shown in \cite{Khaligh:PLOS} that if an object representation is monkey IT-like, it can give a good object recognition performance. 
%Although it is found in \cite{DongWH18} that 
%the response statistics to image object stimuli and intrinsic dimensionality of object representation in DNNs are quite different from those in primate IT cortex as reported in 
%\cite{Lehky2014,Lehky2011}, 
Hence, a further question naturally comes up: \textbf{is the face representation mechanism in CNNs is similar to that in monkey IT cortex found recently in \cite{Chang2017}? or more specifically, could the responses of CNN neurons (units) to face stimuli be linearly modelled in a  parameter space?}
 If so, it would mean that although CNNs generally concatenate multiple convolution layers and nonlinear operators, there essentially exists a linear mapping  between  the face vectors  in a  parameter space and the corresponding face representations in DNNs. This linear mapping is more explicit and largely different from the aforementioned complex image feature encoding
 on CNNs' face representation.

Addressing the above questions, we investigated the face representation problem at higher CNN layers by formulating face images as points in a parameter space in this work, with
six typical multi-layered CNNs for face recognition: VGG-Face \cite{Parkhi15}, DeepID \cite{sun2014deep},  ResNet-Face (defived from ResNet \cite{He7780459}), SphereFace \cite{Liu_2017_CVPR}, Light-CNN \cite{wu2018light}, and ArcFace \cite{deng2018arcface}, and three commonly used face datasets: Multi-PIE \cite{gross2010multi}, LFW \cite{LFWTech}, and MegaFace \cite{kemelmacher2016megaface}. We found that there indeed exists a linear encoding/decoding model for face representation in these CNNs, i.e., face vectors in the  parameter space could not only be effectively decoded from the neuron responses at the higher CNN layers, but also be  encoded linearly for predicting the responses of CNN neurons, similar to the face representation of monkey IT neurons reported in \cite{Chang2017}. In addition, we found that the  predicted representations by  the linear  model could achieve comparable  performances on face recognition and verification to those by the above CNNs.
However, we found that the neuron responses at the higher CNN layers could not be adequately modelled by the axis model in \cite{Chang2017}.
These results partially reveal the linear face representation mechanism in CNNs,
as well as the similarities and differences of face representation between CNNs and primate IT cortex. Additionally, the revealed linear face encoding might also be referenced for the future design of new  face recognition systems.

\section{Method}

\begin{figure}[t]
\centering
  \subfigure {  \includegraphics[width=15 cm]{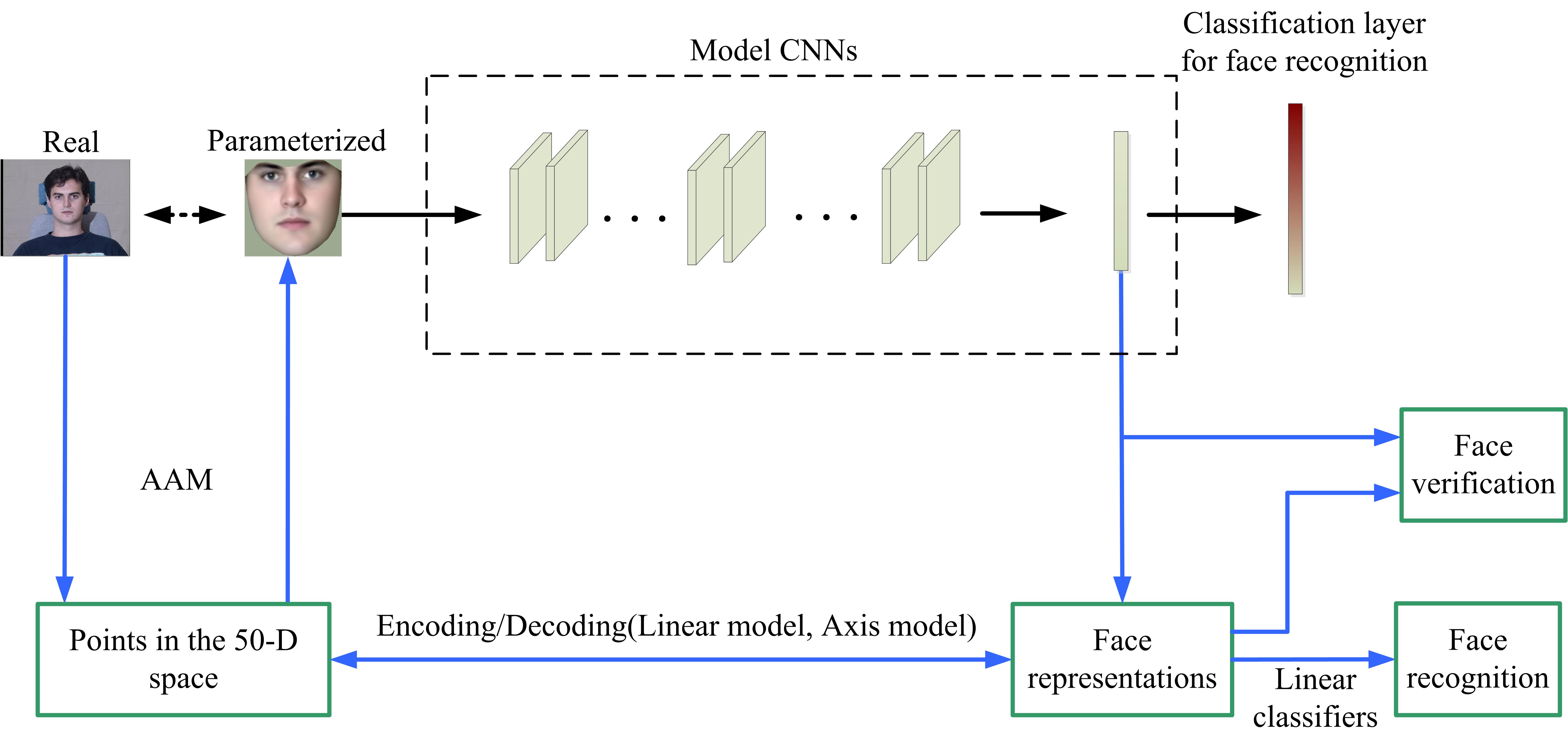}   }
\caption{
Flowchart of our method.} \label{vggface}
\end{figure}

In this section, the used method for investigating the face representation mechanism of CNNs is described, and its flowchart  is shown in Figure \ref{vggface}. The face representation at a CNN layer is defined as:
\begin{definition} \label{def1}
The set of neuron responses at a given layer of a face recognition CNN is defined as the face representation of this layer.
\end{definition}

As seen from Figure \ref{vggface}, we generate the parameterized face images from a given face image dataset using the AAM  (Active Appearance Model) approach \cite{Cootes2001}, and formulate these images as points in a $50$-D(dimensional) parameter space. Then, we analyze the encoding/decoding relationship between the face representations of higher CNN layers and the $50$-D face vectors in the parameter space using a linear model and the proposed axis model in \cite{Chang2017}. In addition, 
considering that face recognition and verification are strongly linked to face representation, we also perform comparative experiments on face recognition and verification with the predicted responses by both the linear model and the axis model.  The details are elaborated in the next.

\subsection{Model CNNs and CNN layer selection for  face representation}

In our work,  the following six popular deep neural networks for face recognition and verification are used 
as our model CNNs:

\noindent \textbf{VGG-Face \cite{Parkhi15}:} It is a typical CNN model for face recognition, derived  from the classical VGG model \cite{Simonyan15} for general object categorization.  It consists of $13$ convolutional layers and $2$ fully connected layers (except the final classification layer for predicting identities).

\noindent \textbf{DeepID \cite{sun2014deep}:} It is a classical CNN model for face verification, aiming to learn so-called deep hidden identity features from face images.  It consists of $4$ convolutional layers  and $1$ fully connected layers.

\noindent \textbf{ResNet-Face:} It is used for face recognition  and verification, derived  from the popular ResNet model  for general object categorization \cite{He7780459}. The code for this model has been  released in the Dlib toolkit\footnote{Dlib toolkit could be downloaded at \url{http://dlib.net/}}.

\noindent \textbf{SphereFace \cite{Liu_2017_CVPR}:} It is a state-of-the-art model for face recognition  and verification based on ResNet, where the angular softmax loss is utilized for learning discriminative face features  with angular margin. In this work, the used model consists of $20$ convolutional layers and $1$ fully connected layers.

\noindent \textbf{Light-CNN \cite{wu2018light}:}
The Light CNN framework  \cite{wu2018light} is to learn a compact embedding on the large-scale face data with massive noisy labels. Here, the Light CNN-29 model, which is a 29-layer convolutional network derived from the Light CNN framework, is utilized.

\noindent \textbf{ArcFace \cite{deng2018arcface}:}
It is a state-of-the-art model for face recognition and verification based on ResNet, where a geometrically interpretable loss function  is utilized. Here, the used model consists of $18$ convolutional layers and $1$ fully connected layers, and the corresponding code is obtained at GitHub \footnote{\url{https://github.com/ronghuaiyang/arcface-pytorch}}.

Considering that higher CNN layers  could generally learn global object information from object stimuli, we investigate the face representations of the neuron responses at Layers $\{13, 14, 15\}$ in VGG-Face, and those at the last fully connected layer (rather than the final classification layer for predicting identities) in DeepID, ResNet-Face, SphereFace, Light-CNN, and ArcFace.

\subsection{Face image synthesis in parameter space} \label{imgsyn}

Although the size of a real face image is usually of millions of dimensions or even higher, it is generally believed that face data lies  on an embedded low-dimensional manifold within the original high-dimensional space \cite{Roweis2000,Tenenbaum2000}.
In order to alleviate the disturbances of the unrelated information to face identity (e.g. background, hair, neck, etc.) in the original high-dimensional form of face data, and simultaneously to reduce the possible information loss  due to the transformation from the  high-dimensional face space to a low-dimensional space, similarly as that in \cite{Chang2017}, we utilize the AAM  approach \cite{Cootes2001} to extract the low-dimensional shape and appearance features of faces from the original face images, and then generate the parameterized face images with these face features  for investigating  the face representation mechanism of CNNs, as shown in Figure \ref{vggface}.

In detail, given a face image dataset, a set of $68$ $2$-D landmark points are automatically extracted from each face image using the Dlib toolkit at first.
Then, the obtained sets of landmark points for all the images are aligned into a common co-ordinate frame and stored as a shape matrix, whose each column represents a aligned set of landmark points extracted from a face image. 
In addition,  the original face images are warped such that its landmark points match the mean shape, and the gray information over the warped region covered by the mean shape is stored as an appearance matrix, whose each column represents the appearance of a warped face image.
Then, Principal Components Analysis (PCA) is applied to the shape matrix for extracting a set of $25$-D feature vectors accounting for the face geometry, and to the appearance matrix for extracting a set of  $25$-D feature vectors accounting for the face appearance. For a given face image, its $25$-D shape  vector and  $25$-D appearance vector are concatenated to form  \textbf{its $50$-D face vector} in this work.

Accordingly, a $50$-D parameter space is spanned by the obtained face vectors, where a point represents a face.
Finally, a parameterized face image is generated with its $50$-D face vector as well as the stored shape and appearance transformation matrices via PCA. 
In our experiments, the obtained face  vectors are used for analyzing the face representation mechanism of CNNs.
The parameterized face images are used for training and testing CNNs.

\noindent \textbf{Remark:}
(i) As described above, compared with the original face images, the parameterized face images have a much less amount of identity-unrelated information to faces (e.g. they do not have complex backgrounds, hairs, and necks), and there is no  information loss generated by the transformation from the high-dimensional parameterized face space to the lwo-dimensional parameter space. Hence,  the parameterized face images seem more appropriate to control and make a strict experimental evaluation on  the face representation mechanism of CNNs than the original ones. 
(ii) Other than  AAM, there exist many other face synthesis approaches in literatures. Here we utilize  AAM  in this work only for conveniently  making a comparison with the  results in \cite{Chang2017} on face representation of IT neurons, where AAM was also utilized.

\subsection{Linear model for face encoding/decoding} 

%\noindent \textbf{Face encoding and decoding via linear regression:}
%To verify whether the face representations (neuron responses at a CNN layer) could be predicted by linearly encoding the face features (also whether the face features could be decoded from the face representations), the transformation
%from neuron responses to feature values (also from feature values to neuron responses) is computed via linear regression. 
%Similarly, 
%
%In addition, once CNN has a linear encoding and decoding representation mechanism, to verify whether  such a mechanism could be modelled by a similar model to the axis model in primate IT cortex for face representation.
 Let $n$ denote the number of face stimuli, and $m$ the number of neurons at a CNN layer. Let $A$ denote the response matrix $R \in \mathcal{R}^{m\times n}$ to all the face stimuli at a CNN layer, and  $P\in \mathcal{R}^{50\times n}$ the matrix for storing the corresponding face vectors defined in the $50$-D parameter space.
A linear model for face encoding and decoding is defined as:
\begin{definition} \label{def2}
Under a linear model for face encoding and decoding,  face encoding could be achieved by linearly transforming a face vector  into the face representation (neuron responses) of a CNN layer, and  face decoding could be achieved by linearly transforming the face representation of a CNN layer into the face vector. 
\end{definition}
 
 If  such a linear model holds true for a CNN on face recognition, the response matrix $R$  for a CNN layer can be roughly approximated by a linear combination of the elements of the $50$-D face vectors $P$ as follows:
\begin{align}
R = TP + b\mathbf{1}_n^T \label{linearreg}
\end{align}
where $T\in \mathcal{R}^{m\times 50}$ is the transformation matrix, $b\in \mathcal{R}^{m\times 1}$ is the bias vector, and $\mathbf{1}_n \in \mathcal{R}^{n\times 1}$ is the $n$-D all-one column vector.

Once both the transformation matrix $T$ and bias vector $b$ are obtained  by solving Eq. (\ref{linearreg}),
the Pearson and Spearman correlation coefficients are computed respectively to measure the  correlation between the neuron responses outputted from a  CNN layer and those predicted by the linear model.

If the  mean of the computed correlation coefficients  is high, it suggests that the face representations of this layer  could be adequately predicted  by linearly encoding the face vectors,
and the face vectors could be linearly decoded
 from the face representations by 
inverting (\ref{linearreg}) accordingly. Otherwise, it suggests that  the face representations of this layer could not be linearly encoded/decoded.

\subsection{The axis model for face encoding/encoding}

The axis model  \cite{Chang2017} can be considered as a special linear model followed by a nonlinear rectification. The axis model consists of two steps: firstly, the
dot product between a face image stimulus (described as a face vector in the  parameter space) and the STA $P_{STA}$ (spike-triggered
average) axis of a face cell is computed, and then the value is rectified  
by a $3$-order polynomial.
%, and the results in \cite{Chang2017} showed that the face representation in primate IT cortex
%could be well modelled by the axis model (also the linear model). 
Here, for a CNN neuron, like that in \cite{Chang2017}, we firstly compute its STA $P_{STA}$ by:
\begin{align}
P_{STA} = \frac{\sum_{i=1}^n r_iP_i}{\sum_{i=1}^n r_i}
\end{align}
where $r_i (i=1,2,...,n)$ is the response of this neuron to the $i$-th face image stimulus, and $P_i$ is the $50$-D face vector of this stimulus. Then, we fit a $3$-order polynomial on the
dot product between the face vector $P_i$ and the STA axis $P_{STA}$ of this neuron for modelling its response $r_i$ by:
\begin{align}
r_i = a + b\langle P_i, P_{STA}\rangle + c\langle P_i, P_{STA} \rangle^2 + d\langle P_i, P_{STA} \rangle^3, \ \ i=1,2,...,n
\end{align}
where $\{a,b,c,d\}$ are the polynomial parameters, and $\langle \cdot, \cdot \rangle$ is the dot product operator.

With the obtained fitted parameters for each CNN neuron, its response  to an arbitrary face image could be predicted, and the Pearson and Spearman correlation coefficients are computed respectively to measure the  correlation between the neuron responses outputted from a CNN layer and those predicted by the axis model.
If the mean of the computed  correlation coefficients is high, it suggests that the axis model could well model the neuron responses at this layer, and the face vectors could also be decoded from the neuron responses with the fitted parameters of the axis model accordingly. Otherwise, it suggests that  the axis model is not appropriate for encoding and decoding the CNN neuron responses at this layer.

\subsection{Face recognition and verification} \label{frv}

Face recognition is to determine the identity of the person in the input face image. It is a multi-class classification problem. For a new face dataset, the original model CNNs (at least their final classification layer) generally have to be fine-tuned  with part of this dataset so that these CNNs would be able to recognize persons from this dataset.

Face verification is to determine whether the persons in
 the input pair of
face images are the same or not. Unlike face recognition, face verification is  typically a binary classification problem, and it does not require fine-tuning the used model CNNs, which could reflect the representation capability of CNNs more generally.

Face representation is the base for face verification and recognition. 
If the neuron responses of higher CNN layers (particularly the last layer) could be adequately predicted by a linear model in the parameter space, 
the predicted responses would achieve similar performances on face verification and recognition
to those outputted from the original model CNNs. Hence, the verification and recognition results could be used indirectly to show the goodness of the predicted face representation.
In this work, we also follow this path to assess the fitness of the linear encoding model, and
the methods used for face recognition and verification are described next:

\noindent \textbf{Remark:} Other than face recognition and verification, we also carried out experiments on face identification, which is to determine the image of a person in a set of face images, who is the same person in the input face image. Our results show that the predicted responses by the linear model achieve similar performances on face identification to those outputted from the original model CNNs, but the predicted responses by the axis model achieve much lower performances than those outputted from the model CNNs, which is in agreement with our results on face recognition and verification. We donot report them in detail, due to the limitation of space.

\noindent \textbf{Face recognition:}

For each of the three used datasets in this work, it is divided into two parts: training data and testing data. We fine-tune a CNN in the following two ways: (i) All the layers of the CNN  are fine-tuned with the training data; (ii) Only the final classification layer is fine-tuned with the training data, while the other layers are fixed, in order to maintain the representation generality of the CNN. Then, the classification accuracies on the testing data are computed.

In addition, we train linear classifiers under two popular loss functions (Softmax Loss and Hinge Loss), with the predicted responses to the training data by the linear model and the axis model respectively, and then compare their performances on the testing data with those of the model CNNs.

The used Softmax-Loss function, by combining the standard Softmax Loss and a regularizer for the loss function, is defined as
\begin{align}
\min_{\theta} \quad -\frac{1}{n}\left[\sum_{i=1}^n\sum_{j=1}^k \mathrm{1}\{y_i=j\} \log\frac{e^{\theta_{j}^Tx_i}}{\sum_{l=1}^k e^{\theta_{l}^Tx_i}} \right] + \frac{\lambda}{2}||\theta||_F^2 
\end{align}
where $n$ is the number of stimuli, $k$ is the number of identities, $x_i \in \mathcal{R}^p$ is the $i$-th input stimulus, $\theta \in \mathcal{R}^{p\times k}$ is the model parameter matrix, 
$y_i\in \{1,2,...,k\}$ is the identity of the $i$-th face image, $\lambda$ is the weight of the regularizer, and  $\mathrm{1}\{\cdot\}$ is the indicator function with $\mathrm{1}\{\mathrm{a \ true \ statement}\} = 1$ and $\mathrm{1}\{\mathrm{a \ false \ statement}\} = 0$.

The used Hinge-Loss function also combines the standard Hinge Loss with a regularizer as:
\begin{align}
\min_{\theta} \ \ -\frac{1}{n}\left[\sum_{i=1}^n \max(0, 1 - \theta_{y_i}^Tx_i + \max_{j\neq y_i}(\theta_j^Tx_i)) \right] + \lambda||\theta||_F^2 
\end{align}

\noindent \textbf{Face verification:}

For the  linear model (also the axis model and the model CNNs),  the verification on  a given  pair of images is carried out  by testing whether the Euclidean distance between the predicted response vectors to the two images is smaller than a threshold $\tau$. 
And the two common measures, Verification Accuracy $Acc$ and Equal Error Rate $EER$, are used for comparing the verification results of the linear/axis model with those of the model CNNs.
 
The Verification Accuracy $Acc$ is defined as follows, and the threshold $\tau$ is generally learned to maximize the verification accuracy on
the training data:
\begin{definition} \label{def3}
$Acc$ is the proportion of true results (both true positives and true negatives) among the total number of cases examined. 
\end{definition}

The Equal Error Rate $EER$ is defined as:
\begin{definition} \label{def4}
$EER$ is  the rate at the ROC (receiver operating characteristic curve) operating point where the false positive and false negative rates are equal. 
\end{definition}
\noindent Note that a smaller value of $EER$
corresponds to a better result, but for comparison convenience,
we report the value of $100\% - EER$ as done in \cite{Parkhi15}.  This measure is independent on the distance
threshold $\tau$.

\section{Results}

\subsection{Data sets}

The following three widely-used face datasets are used in our experiments:

\noindent \textbf{Multi-PIE \cite{gross2010multi}:} It is a popular dataset for algorithmic evaluation on face recognition, containing images of $337$ people with different poses, illuminations, and expressions.  
In our experiments, a subset of Multi-PIE, consisting of the  images of $249$ people
under all the $7$ poses ($\{-45^{\circ},-30^{\circ},-15^{\circ},0^{\circ},+15^{\circ},+30^{\circ},+45^{\circ}\}$) and $10$ illuminations with the neutral
expression in Session One of Multi-PIE (totally $249 \times 7\times 10 = 17430$ images), is used for testing the model CNNs.

\noindent \textbf{LFW \cite{LFWTech}:} It is a standard in-the-wild benchmark for automatic face verification, containing $13233$  images from $5749$ different identities, with large variations in pose, expression and illuminations. 
Following the standard evaluation protocol defined for the “unrestricted setting” \cite{Liu_2017_CVPR, deng2018arcface}, we
test the model CNNs on $6000$ face pairs ($3000$ matched pairs and $3000$ mismatched pairs).

\noindent \textbf{MegaFace \cite{kemelmacher2016megaface}:} It is a standard in-the-wild  benchmark for face verification,
which contains in-the-wild face photos with unconstrained pose, expression, lighting, and exposure. It includes a probe set and a gallery set. The probe set consists of two existing
datasets: Facescrub \cite{NgW14} and FGNet. The gallery set contains around $1$ million images from $690$K different individuals. Considering that our goal in this work is not to evaluate which model DNN performs best on face recognition and verification, but to evaluate (i) whether the encoding and decoding of the face representations of CNNs could be modelled by the linear/axis model and (ii) whether the  linear/axis model could achieve close performances on face  verification in comparison to the model CNNs,
hence, we choose a subset of MegaFace, consisting of $4000$ images from $80$ identities ($40$ males and $40$ females, $50$ images per identity). Then, we construct $6000$ face pairs ($3000$ matched pairs and $3000$ mismatched pairs) with the subset of images for our experiments.

As described in Section \ref{imgsyn}, the $50$-D face vectors are extracted from the original images in the three datasets  using the AAM approach.
Then, the parameterized face images are generated using these $50$-D face vectors.

Following the common practice, the three synthesized face datasets are partitioned into two subsets separately: a training set and a testing set.  The training set is used for estimating the fitted parameters in both the linear model and the axis model,  fine-tuning VGG-Face and training the linear classifiers for the face recognition experiments, while  the testing set is  only to test the  face representation performances of our linear model as well as the axis model. 
For Multi-PIE, five data partition schemes are assessed in order to give a detailed analysis on the influences of viewing pose and illumination, which are listed in Table \ref{fnet}.
For LFW and MegaFace, the aggregate
performance of each CNN on 10 separate experiments are evaluated in a
10-fold cross validation scheme. In each experiment,
nine of the subsets are combined to form a training
set, with the remaining subset used for testing.

%\noindent \textbf{Remark:} To guarantee the generality of our results, the dataset used in our experiments is divided into two parts: a training set and a testing set. The training set is used for fine-tuning the model CNN (VGG-Face) and learning the AAM model at first, and then both the training set and testing set are  used for analyzing the face representation mechanism of CNN. 

\begin{table}[t]
  \footnotesize
  \centering
  %\begin{tabular}{p{50pt}p{70pt}p{70pt}p{70pt}p{70pt}p{70pt}}
  \begin{tabular}{|c|c|} \hline
    %\toprule
    %\hline
    Index & Partition schemes (for each identity)  \\
    \hline
    1     & Select samples with poses $\{-30^{\circ},-15^{\circ},0^{\circ},45^{\circ}\}$ for training, the rest for testing.  \\
    \hline
    2     & Select samples with poses $\{-30^{\circ},-15^{\circ},0^{\circ},30^{\circ},45^{\circ}\}$ for training, the rest for testing. \\
    \hline
    3     & Select samples with 6 random illuminations for training, the rest for testing.  \\
    \hline
    4     & Select samples with poses $\{-45^{\circ},15^{\circ},30^{\circ}\}$ for training, the rest for testing.  \\
    \hline
    5    & Select samples with 3 random poses for training, the rest for testing. \\
    \hline
  \end{tabular}
  \caption{Partition schemes for constructing the training and testing sets in Multi-PIE.} \label{fnet}
\end{table}

\subsection{Encoding/decoding under the  linear model}  \label{expLinear}

\noindent \textbf{Results on Multi-PIE:}

For this relatively simple dataset, only VGG-Face  \cite{Parkhi15} is tested. 
As described in Section \ref{frv}, each of the five training sets (as described in Table \ref{fnet})  is used to fine-tune VGG-Face in two different ways, and we denote: the model by fine-tuning its classification layer while fixing its rest layers as VGG-Face1 and the model by fine-tuning all the layers as VGG-Face2.

\begin{figure}[t]
\centering
%  \subfigure[] { \includegraphics[width=5 cm]{CorrPearsonPop.jpg} \label{CorrPearsonPop}  }
%  \subfigure[] { \includegraphics[width=5 cm]{CorrSpearmanPop.jpg} \label{CorrSpearmanPop} } \\
  \subfigure[VGG-Face1] {  \includegraphics[width=5.5 cm]{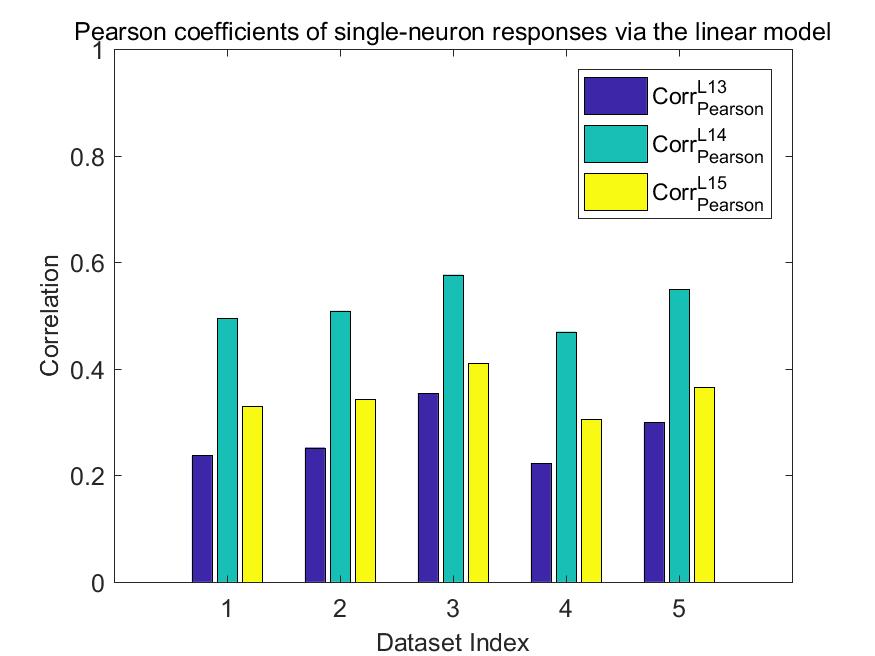} 
  \label{CorrPearsonSingle1}  }
  \subfigure[VGG-Face1] {  \includegraphics[width=5.5 cm]{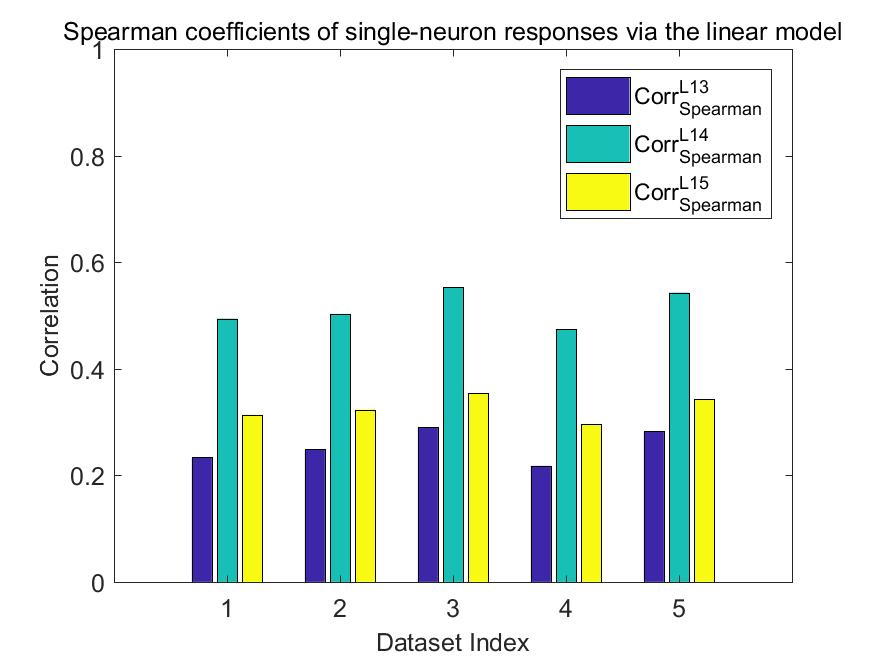} \label{CorrSpearmanSingle1}   
  }
  \subfigure[VGG-Face2] {  \includegraphics[width=5.5 cm]{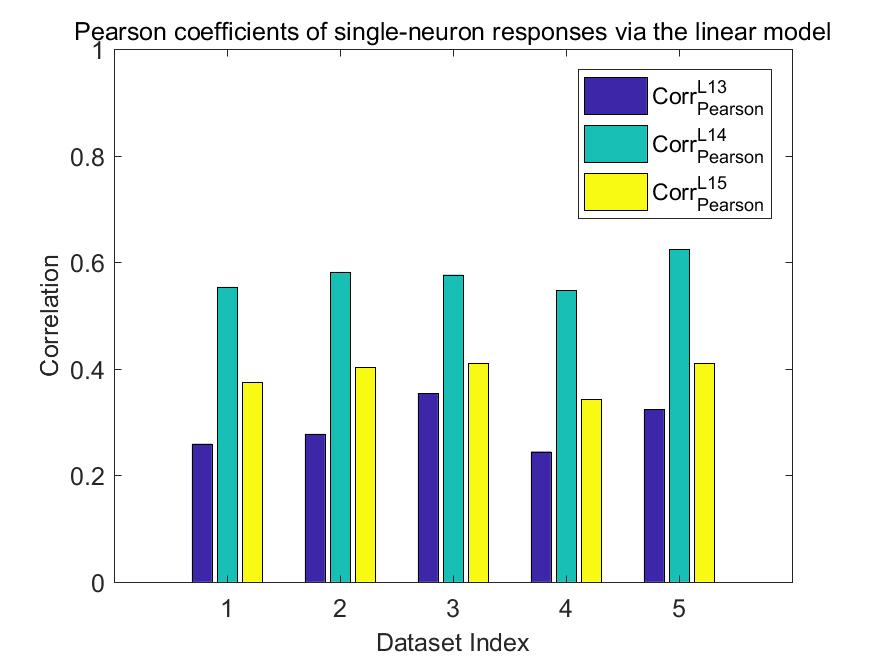} \label{CorrPearsonSingle}  }
  \subfigure[VGG-Face2] {  \includegraphics[width=5.5 cm]{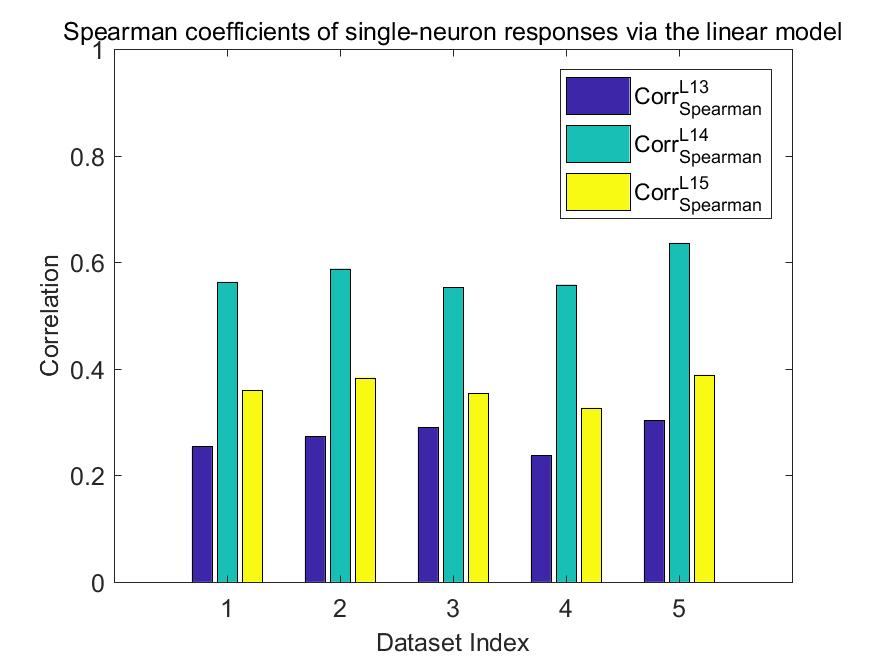} \label{CorrSpearmanSingle}   
  }
  
\caption{
Pearson and Spearman coefficients on Multi-PIE between the single neuron responses from Layers  $\{L13,L14,L15\}$ of VGG-Face1/VGG-Face2 and those by the linear model: (a) Pearson coefficient comparison to VGG-Face1; (b) Spearman coefficient comparison  to VGG-Face1; (c) Pearson coefficient comparison to VGG-Face2; (d) Spearman coefficient comparison  to VGG-Face2.}   \label{corrLG}
\end{figure}

\begin{figure}[t]
\centering
  \subfigure[Multi-PIE] { \includegraphics[width=10 cm]{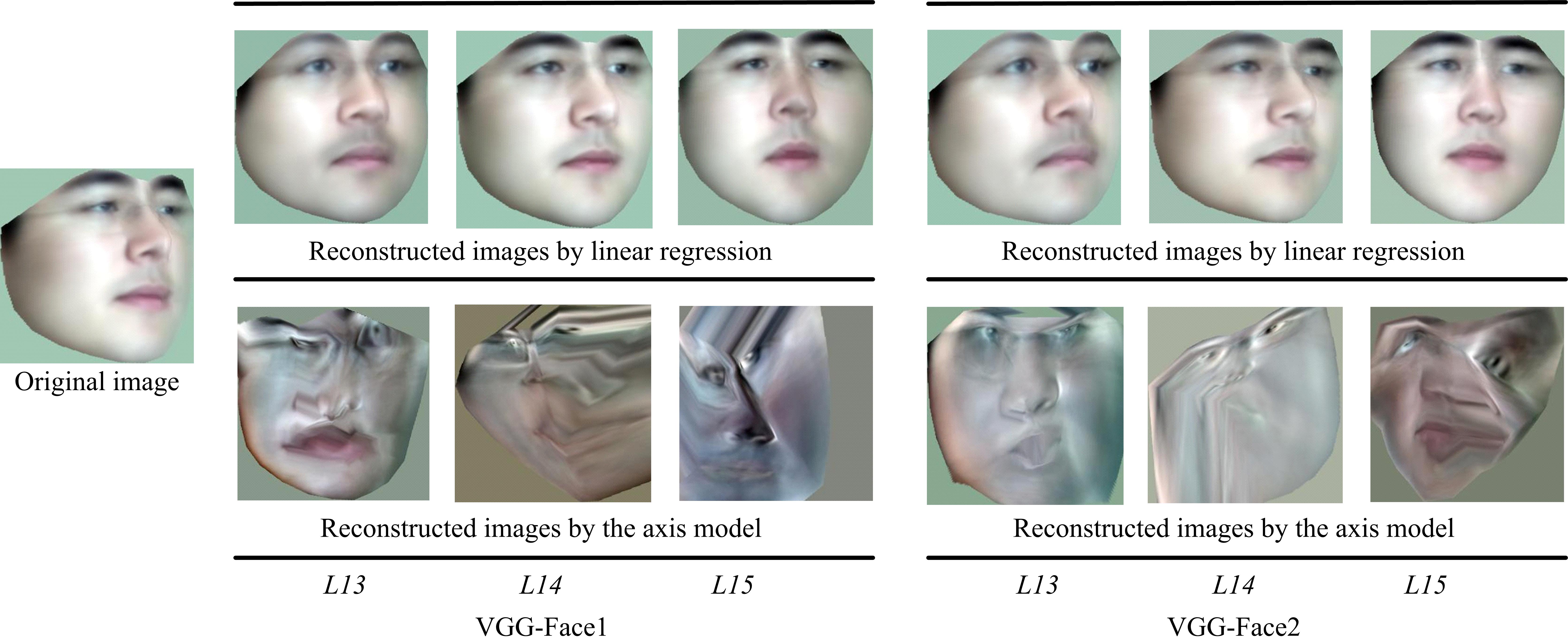} \label{reconPIE} }
  \subfigure[MegaFace] { \includegraphics[width=10 cm]{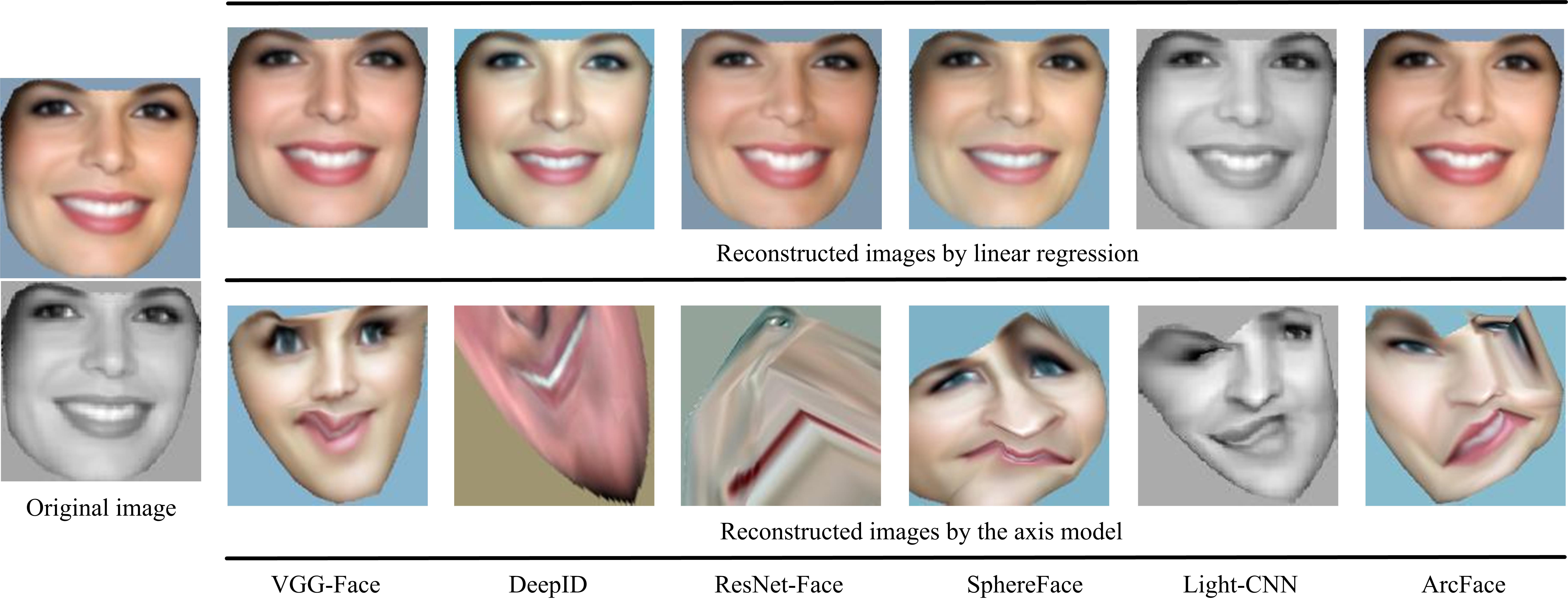} \label{reconMega} }
\caption{
Reconstructed examples by the linear model and the axis model: (a) Reconstructed examples (Multi-PIE) from the responses of VGG-Face1/VGG-Face2; (b) Reconstructed examples (MegaFace) from the responses of all the six CNNs.}   \label{recon}
\end{figure}

Following Eq. (\ref{linearreg}), we fit the linear model  between the $50$-D face vectors of the training data  and the corresponding neuron responses at each of Layers $\{L13,L14,L15\}$ for VGG-Face1 and VGG-Face2 respectively, and the obtained model parameters are used for predicting the neuron responses of the selected three layers to the testing data.
After that, the Pearson and Spearman coefficients are computed between the predicted single neuron responses and those outputted from the three layers, and the mean values of the two coefficients on the five testing sets are shown in Figure \ref{corrLG}.
As seen from Figure \ref{corrLG}, the computed Pearson and Spearman coefficients for $L14$ of both VGG-Face1 and VGG-Face2 are around $0.6$, and the computed Pearson and Spearman coefficients for $L15$ of both VGG-Face1 and VGG-Face2 are close to $0.4$. 
%The relative magnitude order of the two coefficients on the referred sets are consistent  that $L14 > L15 > L13$. 
This suggests that the predicted representations by the linear model  are strongly  correlated with those outputted from Layers $\{L14, L15\}$.

We also linearly decode the $50$-D face vectors from the neuron responses  at Layers  $\{L13,L14,L15\}$ respectively, then reconstruct the synthesized face images according to the AAM approach.
Figure \ref{reconPIE} shows the reconstructed results on an examplar image at Layers $\{L13,L14,L15\}$ of VGG-Face1 and VGG-Face2, and it can be seen that these reconstructed images are similar to the original synthesized face image, indicating that 
the representations outputted by higher CNN layers  could also be utilized for linearly decoding the face vectors in the $50$-D parameter space.

\begin{figure}[t]
\centering
  \subfigure[To VGG-Face1] {  \includegraphics[width=5.5 cm]{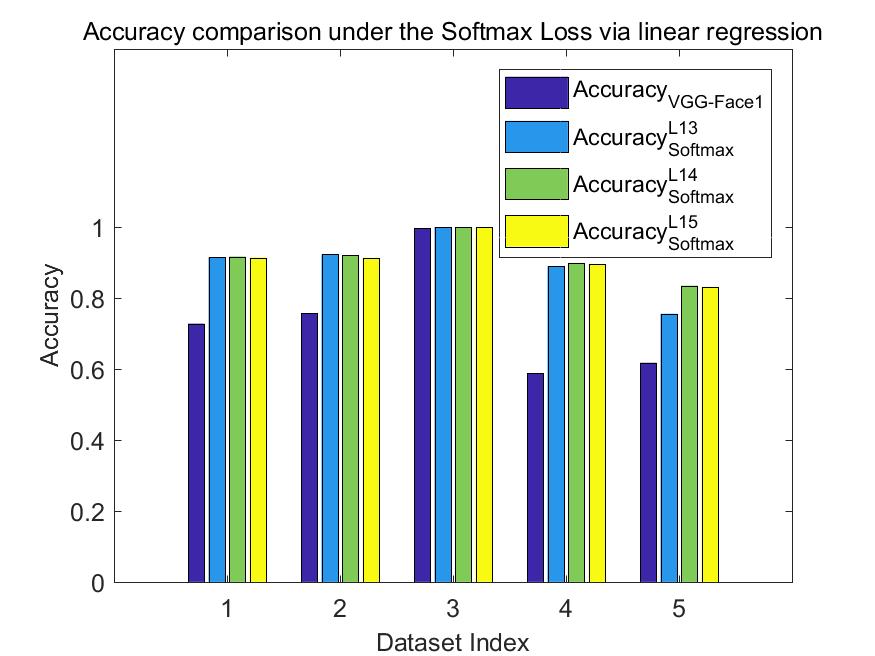} \label{faccsoft}  }
  \subfigure[To VGG-Face1] {  \includegraphics[width=5.5 cm]{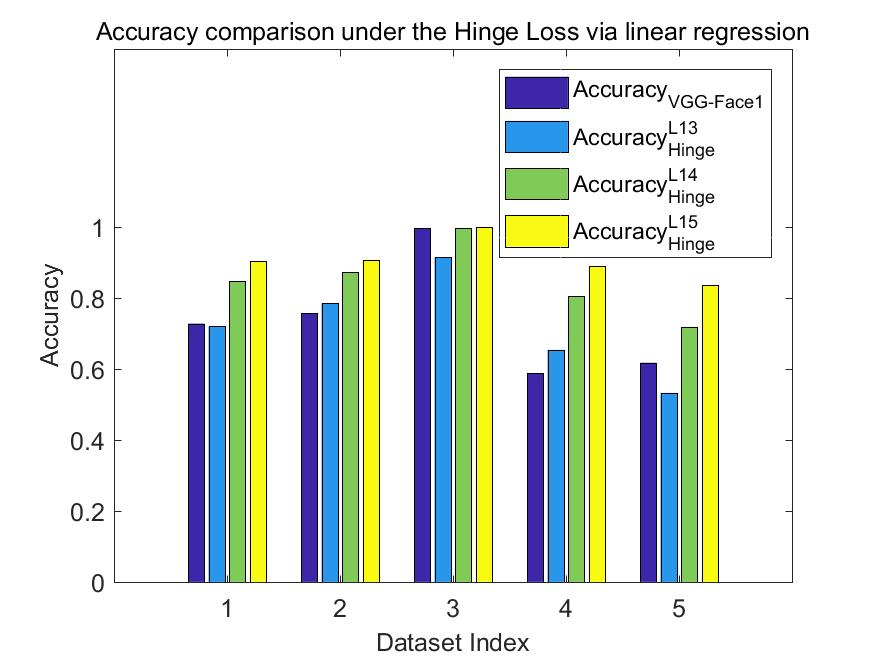} \label{facchinge}  }
  \subfigure[To VGG-Face2] {  \includegraphics[width=5.5 cm]{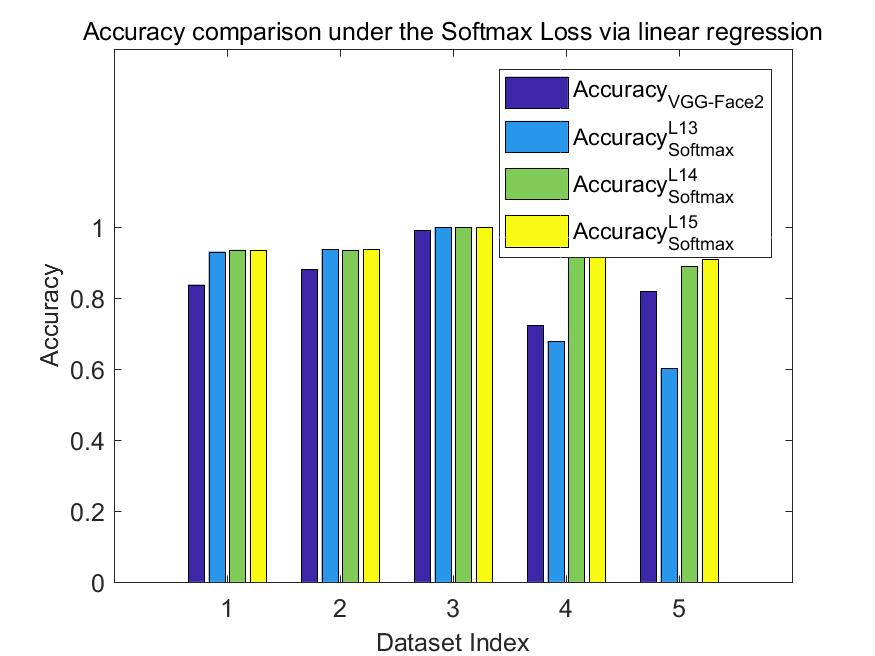} \label{faccsoft2}  }
  \subfigure[To VGG-Face2] {  \includegraphics[width=5.5 cm]{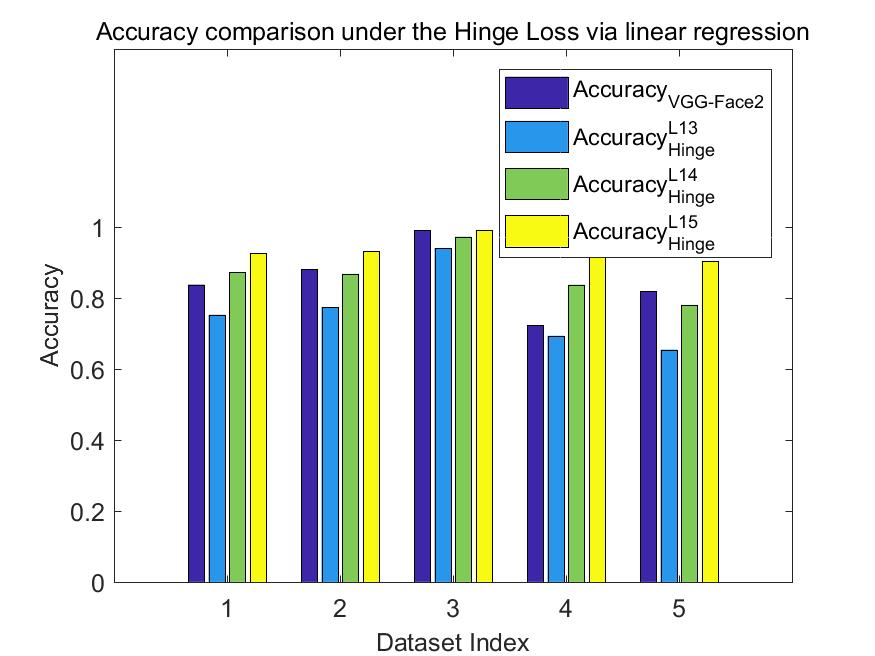} \label{facchinge2}  }
\caption{
Accuracies of VGG-Face1/VGG-Face2 and the linear classifiers that use  the predicted representations for Layers $\{L13,L14,L15\}$ by the linear model: (a) Accuracy comparison under the Softmax Loss to VGG-Face1; (b) Accuracy comparison under the Hinge Loss to VGG-Face1; (c) Accuracy comparison under the Softmax Loss to VGG-Face2; (d) Accuracy comparison under the Hinge Loss to VGG-Face2.}\label{acc}
\end{figure}

In addition, using the predicted representations by the linear model for each of Layers $\{L13,L14,L15\}$ of both VGG-Face1 and  VGG-Face2,
we train the linear classifiers for face recognition under the Softmax Loss and the Hinge Loss respectively. Then, we evaluate their performances  on  the five testing sets, and Figure \ref{acc}
reports the recognition accuracies by VGG-Face1/VGG-Face2 and the linear model.
As seen from the dark blue bars of Figure \ref{acc}, the  recognition accuracies by VGG-Face1 on the five testing sets are $\{72.70\%, 75.74\%, 99.54\%, 58.78\%, 61.71\%\}$, and the recognition accuracies by VGG-Face2 are $\{83.73\%,	88.05\%, 99.01\%, 72.33\%, 81.90\%\}$.
The learnt linear classifiers for Layers $\{L14, L15\}$ achieve close or  better performances than VGG-Face1/VGG-Face2, while the linear classifiers for Layer $L13$  achieve close performances to VGG-Face1/VGG-Face2 in most cases.

Note that VGG-Face2 achieves close or  slightly better performances than VGG-Face1, mainly because VGG-Face2 is obtained by fine-tuning all the layers. And it is also noted that the accuracies on the testing sets with Nos. $\{1,2,4,5\}$ are much lower than those on the third set, mainly because (i) the images in the testing sets with Nos. $\{1,2,4\}$ have different head orientations from those in their corresponding training sets, and (ii) for each identity, its face images in the fifth testing set have different head orientations from those in the corresponding training set.

\noindent \textbf{Results on LFW and MegaFace:}

For the two in-the-wild datasets, all the six CNNs without fine-tuning are used for conducting face verification experiments to further investigate whether their face representations could be adequately modelled by linear encoding.
 
\begin{figure}[t]
\centering

  \subfigure[On LFW] {  \includegraphics[width=5.5 cm]{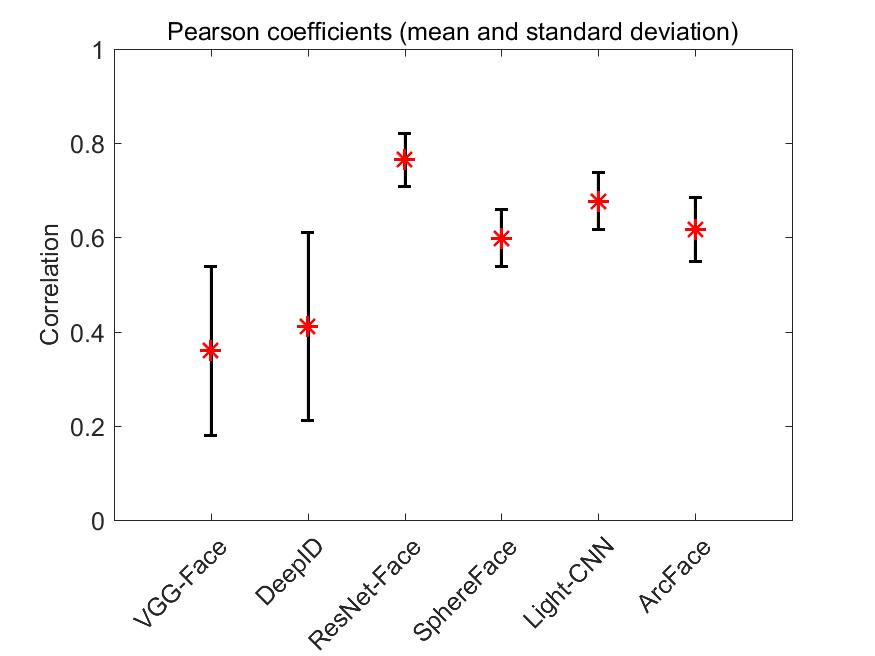} 
  \label{CorrPearsonSingle1}  }
  \subfigure[On LFW] {  \includegraphics[width=5.5 cm]{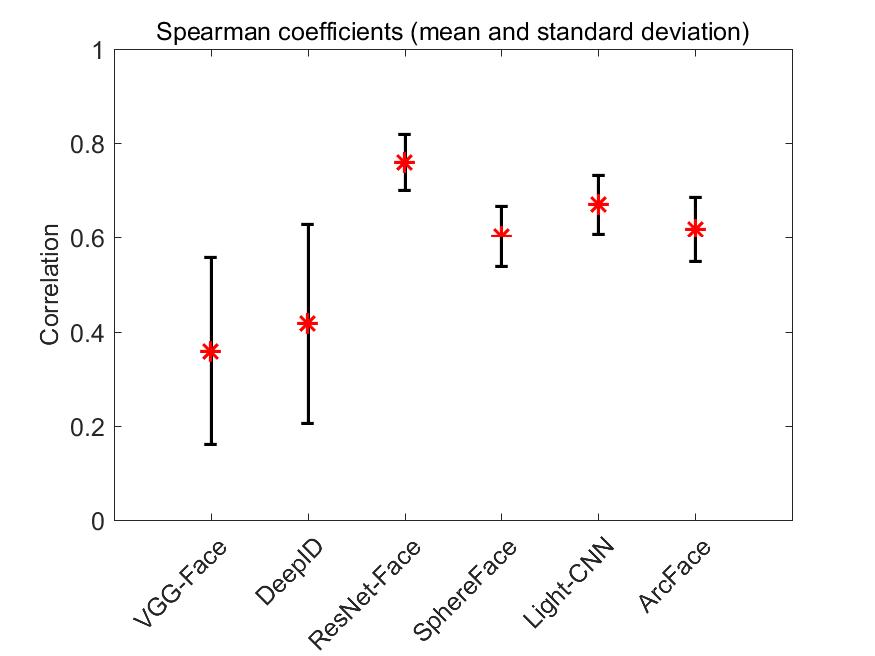} \label{CorrSpearmanSingle1}   
  }
  \subfigure[On MegaFace] {  \includegraphics[width=5.5 cm]{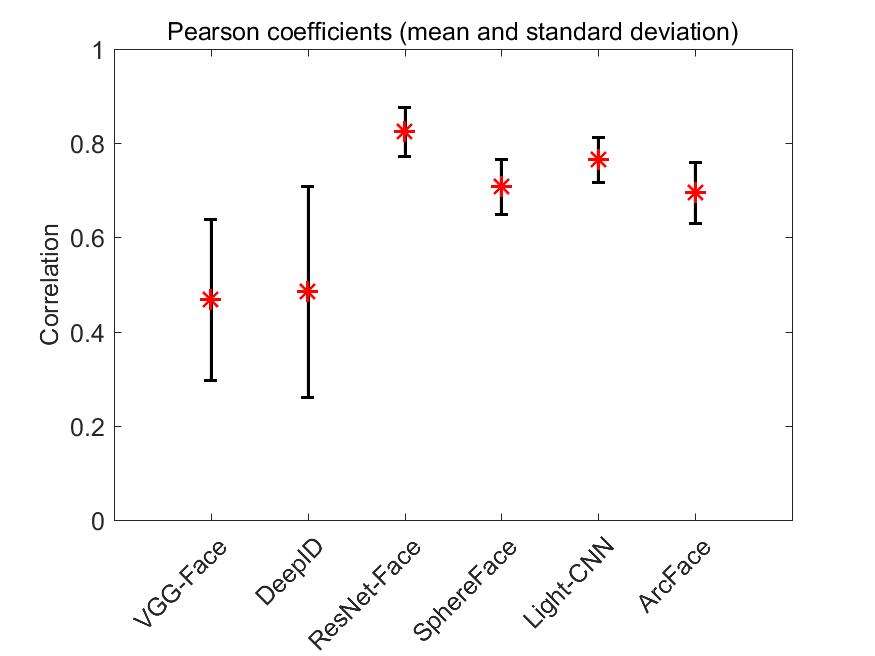} \label{CorrPearsonSingle}  }
  \subfigure[On MegaFace] {  \includegraphics[width=5.5 cm]{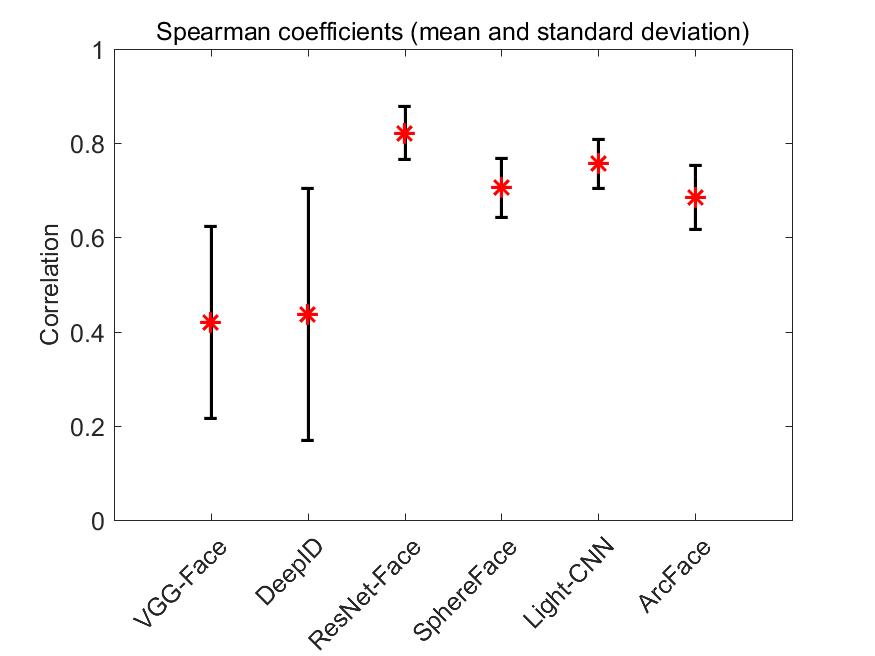} \label{CorrSpearmanSingle}   
  }
\caption{
Pearson and Spearman coefficients between the single neuron responses from all the referred CNNs and those by the linear model: (a) Pearson coefficients on LFW; (b) Spearman coefficients   on LFW; (c) Pearson coefficients on MegaFace; (d) Spearman coefficients  on MegaFace.}   \label{corrLFWMEGA}
\end{figure}

As described in Section \ref{frv}, each CNN on 10 separate experiments are evaluated in a
10-fold cross validation scheme. In each experiment,
we fit the linear model  between the $50$-D face vectors of the training data  and the corresponding neuron responses at the last layer of each referred CNN. Accordingly, the obtained model parameters are used for predicting the neuron responses of this layer to the testing data.
Then, the Pearson and Spearman correlation coefficients are computed between the predicted single neuron responses and those outputted from this layer. 
The significance of the computed correlations is also tested, and more than $94\%$ of the corresponding $p$-values  for each CNN (close to $100\%$ for ResNet-Face, SphereFace, Light-CNN, ArcFace) are lower than the significance level of $0.01$. 
The mean values (also the standard deviations) of the two computed correlation coefficients by all the referred CNNs are shown in Figure \ref{corrLFWMEGA}.
As seen from Figure \ref{corrLFWMEGA}, both the computed Pearson and Spearman coefficients for the six CNNs on the two datasets are larger than $0.4$ in most cases, and particularly, those coefficients for the four more recent CNNs (ArcFace, Light-CNN, SphereFace, and ResNet-Face) are even close to or larger than $0.6$ with relatively smaller standard deviations, in agreement with the previous results for VGG-Face1/VGG-Face2 on Multi-PIE, which further suggests that the predicted representations by the linear model  are strongly  correlated with those outputted from CNNs.

We also linearly decode the $50$-D face vectors from the neuron responses  at the last Layer of each CNN respectively, then reconstruct the synthesized face images according to the AAM approach.
Figure \ref{reconMega} shows the reconstructed results on an examplar image from the in-the-wild dataset MegaFace, and  these reconstructed images are similar to the original synthesized face image (Since Light-CNN uses a grey image as input, its reconstructed image is also a grey image). The results once again indicate that the representations outputted by higher CNN layers could  be utilized for linearly decoding the face vectors in the $50$-D parameter space.

\begin{figure}[t]
\centering

  \subfigure[$ACC$ on LFW] {  \includegraphics[width=5.5 cm]{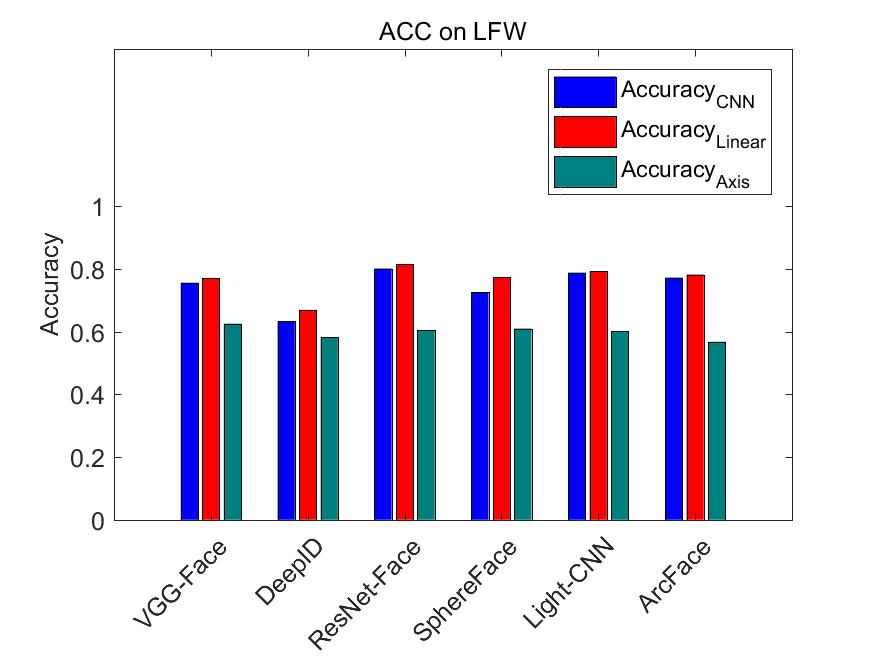} 
  \label{CorrPearsonSingle1}  }
  \subfigure[$100\% - EER$ on LFW] {  \includegraphics[width=5.5 cm]{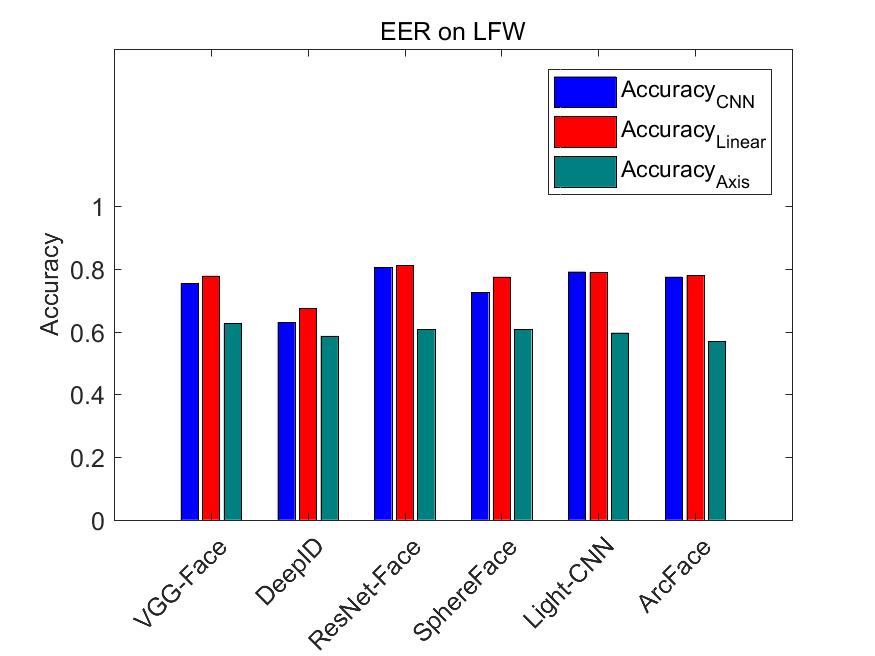} \label{CorrSpearmanSingle1}   
  }
  \subfigure[$ACC$ on MegaFace] {  \includegraphics[width=5.5 cm]{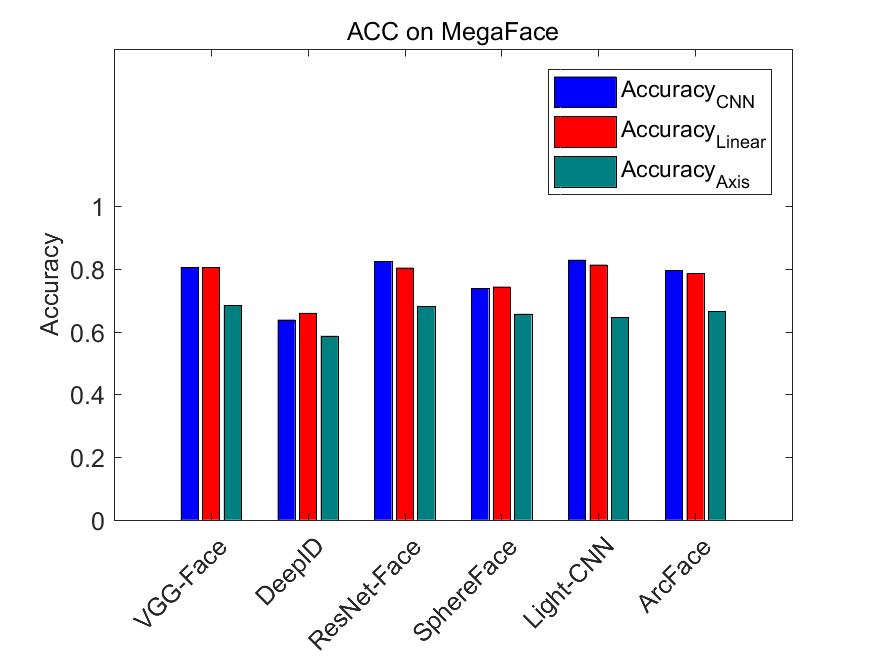} \label{CorrPearsonSingle}  }
  \subfigure[$100\% - EER$ on MegaFace] {  \includegraphics[width=5.5 cm]{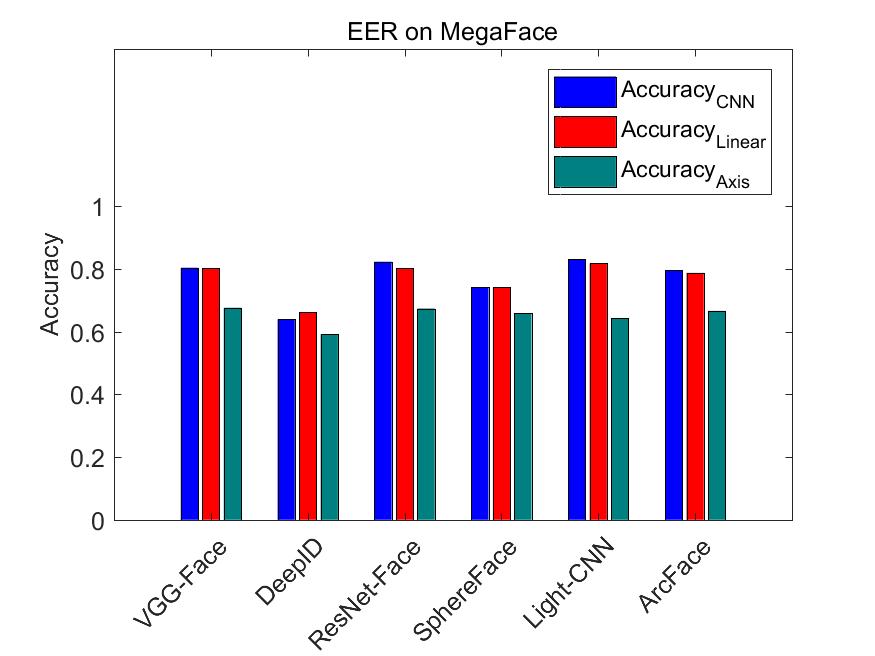} \label{CorrSpearmanSingle}   
  }
\caption{
$ACC$ and $EER$ on LFW and MegaFace with the presentations of the six CNNs and the corresponding predicted presentations by the linear/axis model.}   \label{accLFWMEGA}
\end{figure}

In addition, the experiments on face verification with the representation of each CNN and the  predicted representation by the linear model are conducted, and the corresponding $ACC$ and $EER$ (in face, $100\% - EER$ ) on LFW and MegaFace are shown in Figure \ref{accLFWMEGA}.
The results show that all the predicted representations achieve close performances to the corresponding CNN representations, in agreement with the above results on face recognition.

In sum, all the above results indicate:
\begin{itemize}
\item[1] 
The  representations of higher CNN layers  could be well predicted by the linear  model, and notably, the linear model tends to give a better prediction for the representations of more recent CNNs. 
\item[2] The face vectors in the parameter space could be well decoded from the CNN representations by the linear model. 
%The computed recognition accuracies and correlation coefficients  are in agreement with the assumption in the deep learning  community that the penultimate layer of a CNN (Layer $L15$ of VGG-Face in this work) is able to learn a more discriminative representation by discarding some identity-orthogonal face features  to some extent.
\end{itemize}

\noindent \textbf{Remark:} Similar to the nonlinear rectification used in \cite{Chang2017}, after obtaining the  responses fitted by the linear model, we also tried to rectify these responses with a $3$-order polynomial, and found that such a rectification step did not
affect the encoding/decoding results.

\subsection{Encoding/decoding under the axis model}

\begin{figure}[t]
\centering
  \subfigure[VGG-Face1] {  \includegraphics[width=5.5 cm]{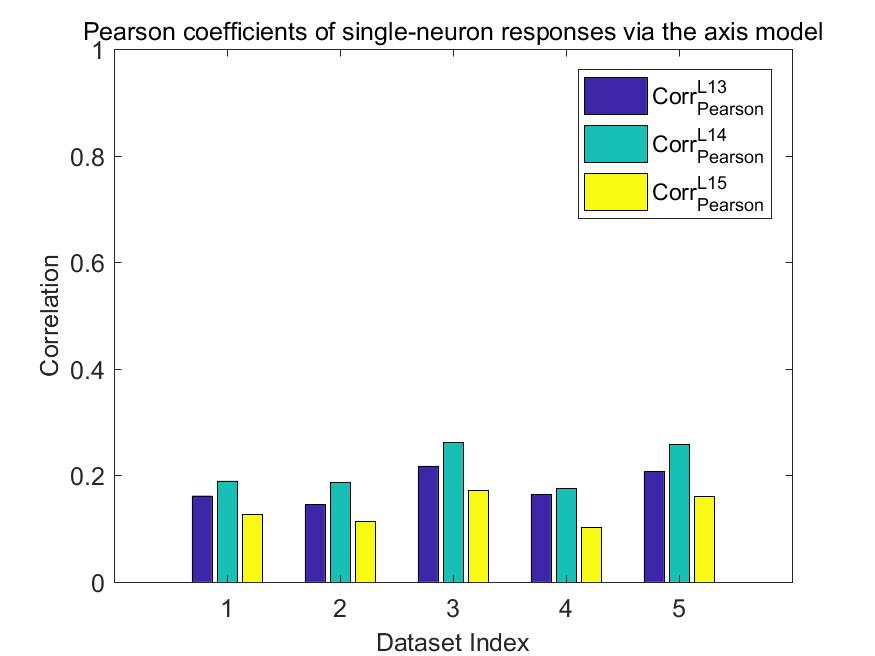} \label{CorrPearsonSingle1}  }
  \subfigure[VGG-Face1] {  \includegraphics[width=5.5 cm]{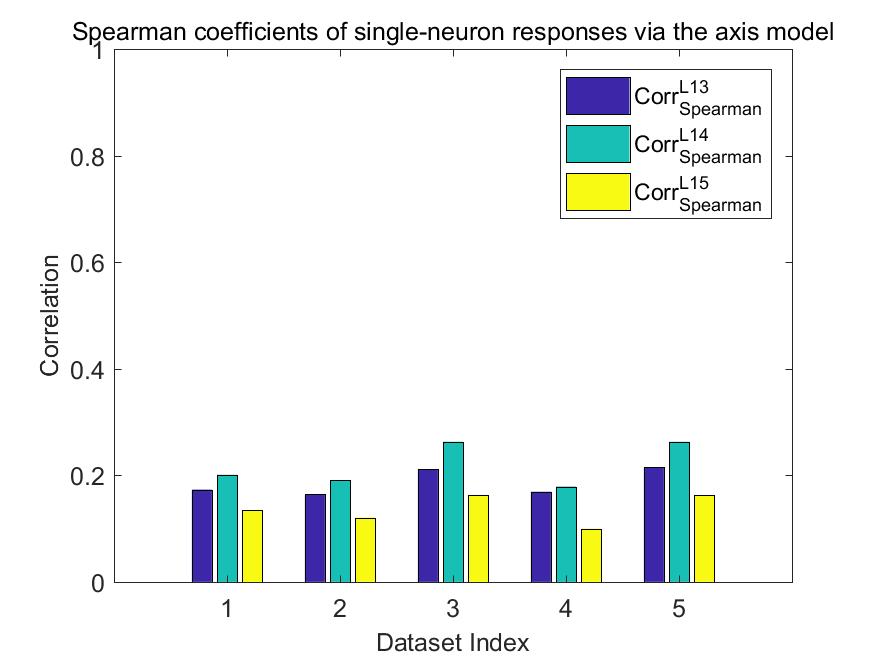} \label{CorrSpearmanSingle1}  }
  \subfigure[VGG-Face2] {  \includegraphics[width=5.5 cm]{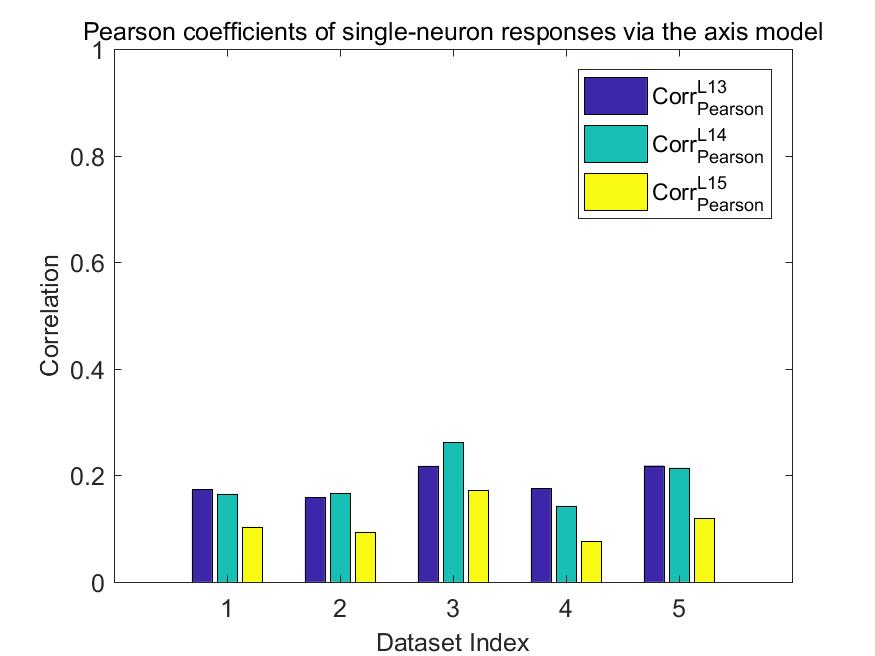} \label{CorrPearsonSingle}  }
  \subfigure[VGG-Face2] {  \includegraphics[width=5.5 cm]{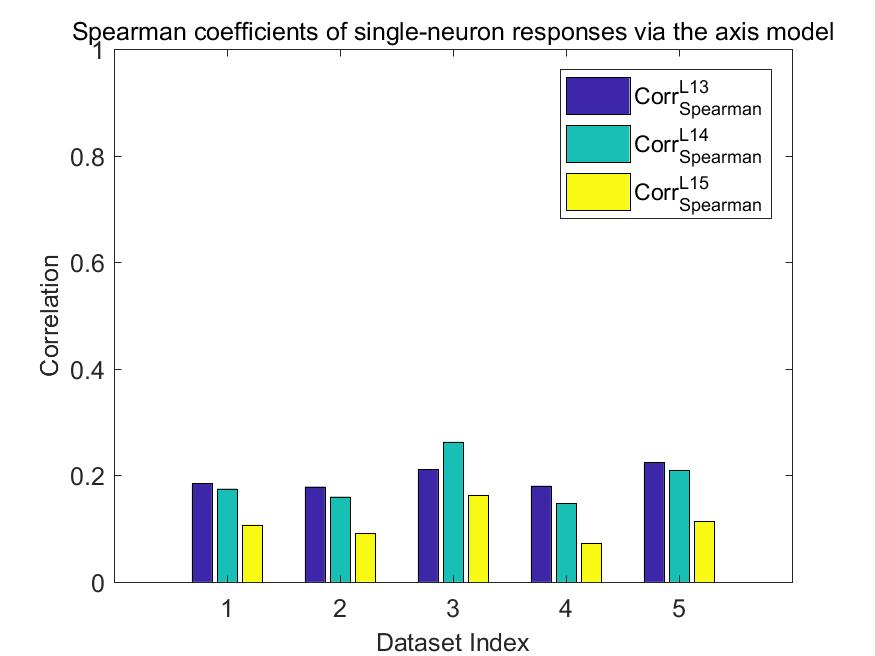} \label{CorrSpearmanSingle}  }
  
\caption{
Pearson and Spearman coefficients on Multi-PIE between the single neuron responses from Layers  $\{L13,L14,L15\}$ of VGG-Face1/VGG-Face2 and those by the axis model: (a) Pearson coefficient comparision to VGG-Face1; (b) Spearman coefficient comparision to  VGG-Face1; (c) Pearson coefficient comparision to  VGG-Face2; (d) Spearman coefficient comparision to  VGG-Face2.}   \label{corrLGAxis}
\end{figure}

In this subsection, we investigate whether the proposed axis model in \cite{Chang2017} for primate IT cortex is suitable for modelling the face representations of higher CNN layers. 

The same procedure in Section \ref{expLinear} is used here, except that the axis model is used to replace the linear model in Section \ref{expLinear}. The results are summarized as follows:
 
\noindent \textbf{Results on Multi-PIE:}

Figure \ref{corrLGAxis} shows the mean values of the Pearson and Spearman correlation coefficients on the five testing sets from Multi-PIE for Layers $\{L13, L14, L15\}$ of VGG-Face1/VGG-Face2. These coefficients are lower than $0.25$ in most cases, indicating that the predicted responses  by the axis model are not  strongly correlated with those outputted from Layers $\{L13, L14, L15\}$ of VGG-Face1 and VGG-Face2.

Figure \ref{reconPIE} shows the reconstructed results on an examplar image at Layers $\{L13,L14$, $L15\}$ by the axis model. The reconstructed images are dramatically different from the original face image, indicating that the axis model cannot effectively decode the face features from the representations outputted from higher CNN layers, although it was successful for IT face neuron decoding in \cite{Chang2017}.

\begin{figure}[t]
\centering
  \subfigure[To VGG-Face1] {  \includegraphics[width=5.5 cm]{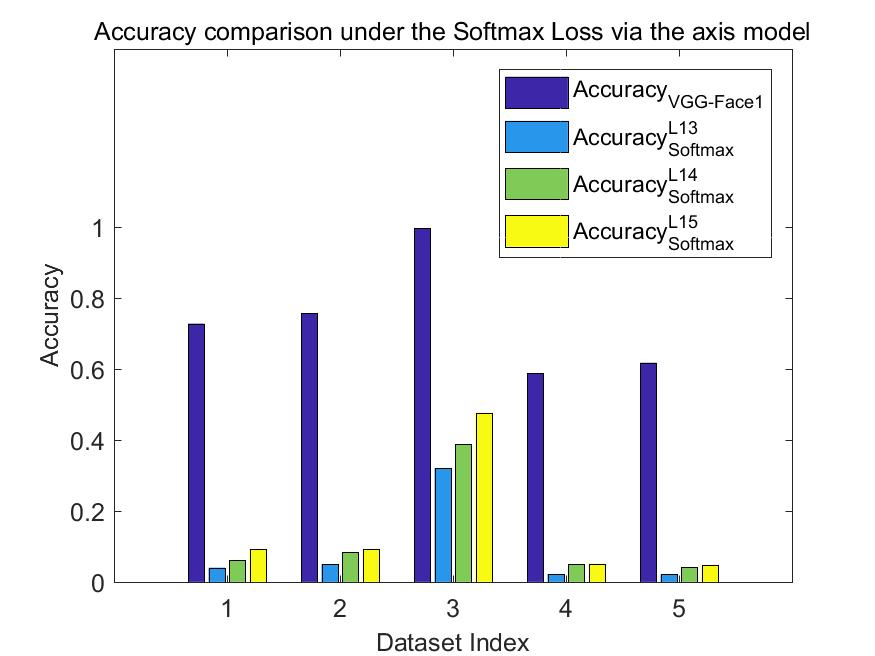} \label{faccsoftAxis1}  }
  \subfigure[To VGG-Face1] {  \includegraphics[width=5.5 cm]{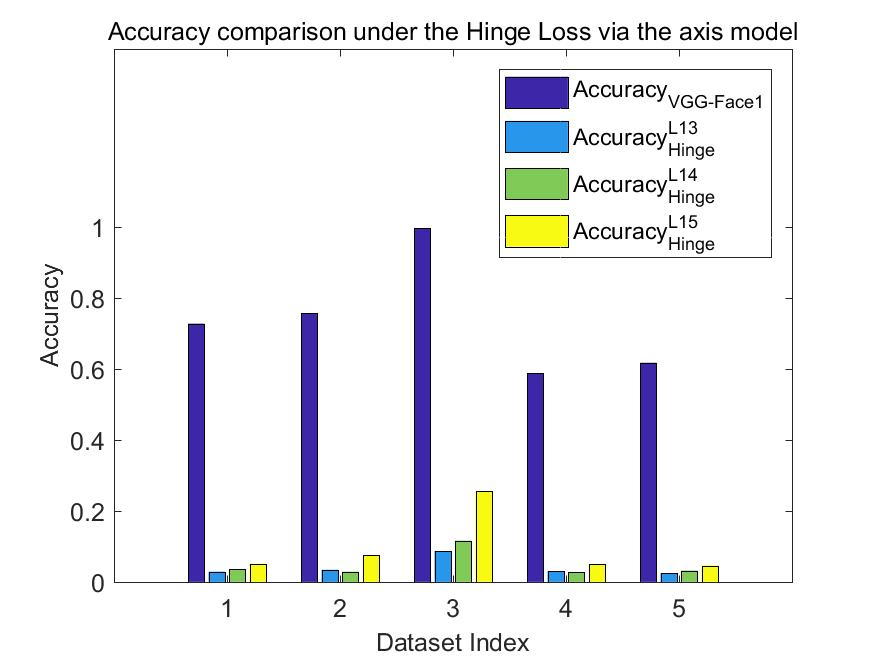} \label{facchingeAxis1}  }
  \subfigure[To VGG-Face2] {  \includegraphics[width=5.5 cm]{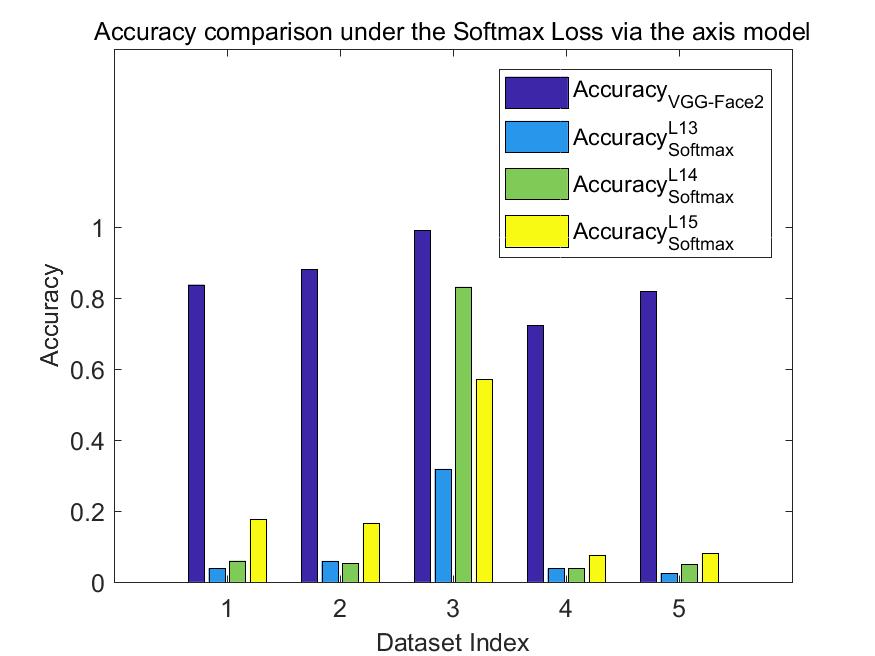} \label{faccsoftAxis}  }
  \subfigure[To VGG-Face2] {  \includegraphics[width=5.5 cm]{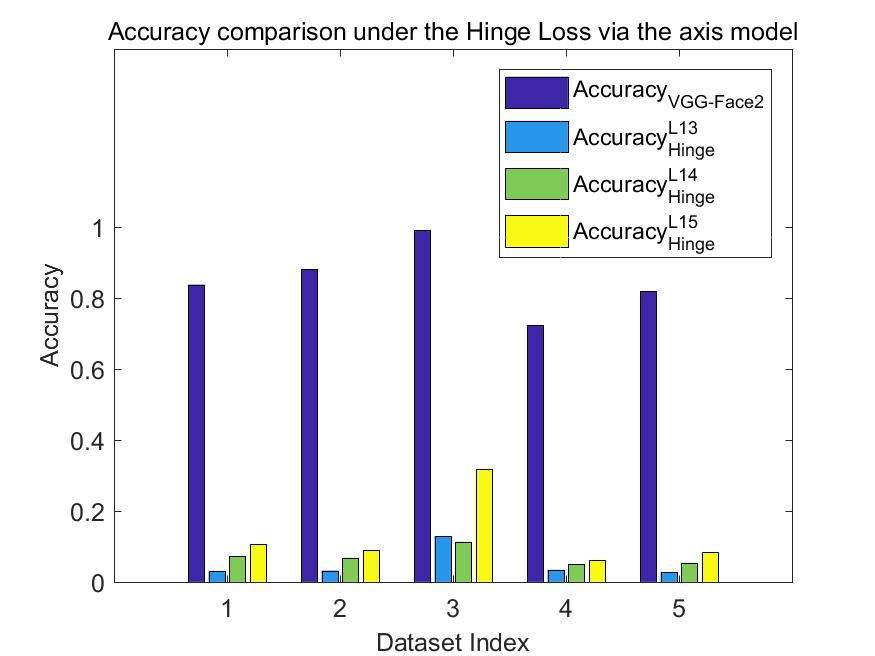} \label{facchingeAxis}  }
\caption{
Accuracies of VGG-Face1/VGG-Face2 and the linear classifiers that use  the predicted representations for Layers $\{L13,L14,L15\}$ by the axis model: (a) Accuracy comparison under the Softmax Loss to VGG-Face1; (b) Accuracy comparison under the Hinge Loss to VGG-Face1; (c) Accuracy comparison under the Softmax Loss to VGG-Face2; (d) Accuracy comparison under the Hinge Loss to VGG-Face2.}\label{accAxis}
\end{figure}

The face recognition accuracies of VGG-Face1/VGG-Face2 and the learned linear classifiers  
from the predicted representations for Layers $\{L13,L14,L15\}$ by the axis model are shown in Figure \ref{accAxis}.
As seen from Figure \ref{accAxis}, the learnt linear classifiers on the third training set perform better than those on the other four training sets, mainly because the third training set contains all the possible poses in the corresponding testing set. However, 
the linear classifiers for the three layers on all the training sets are much less accurate than the corresponding CNN  in most cases, which further demonstrates that the axis model is not suitable for modelling the face representations in CNNs.

\noindent \textbf{Results on LFW and MegaFace:}

\begin{figure}[t]
\centering

  \subfigure[On LFW] {  \includegraphics[width=5.5 cm]{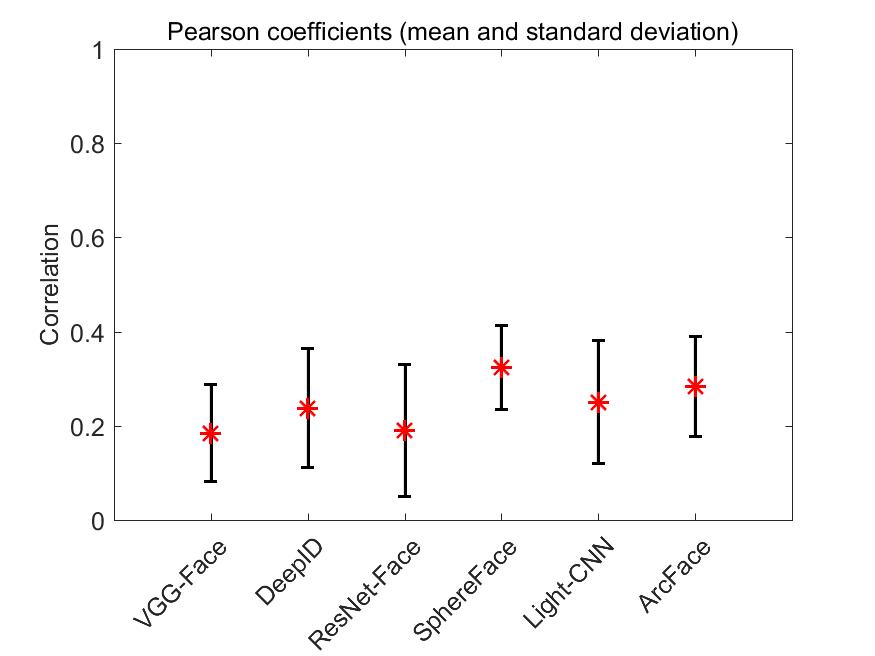}   }
  \subfigure[On LFW] {  \includegraphics[width=5.5 cm]{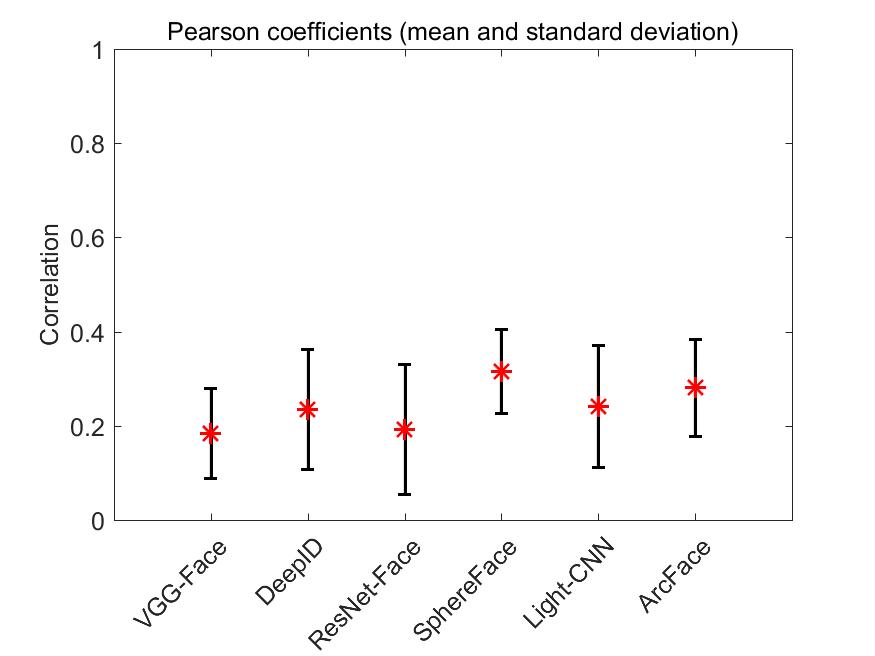}  }
  \subfigure[On MegaFace] {  \includegraphics[width=5.5 cm]{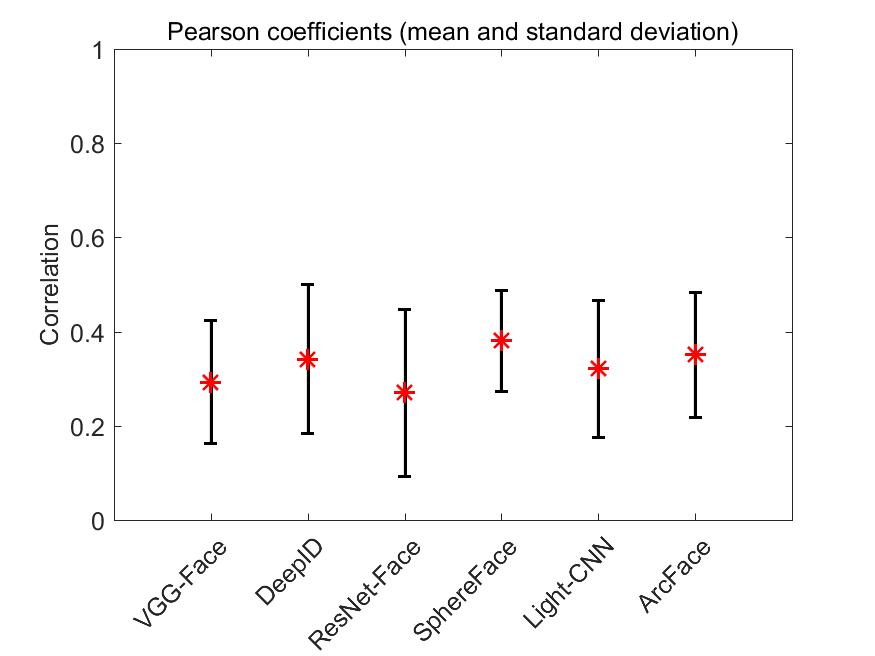}  }
  \subfigure[On MegaFace] {  \includegraphics[width=5.5 cm]{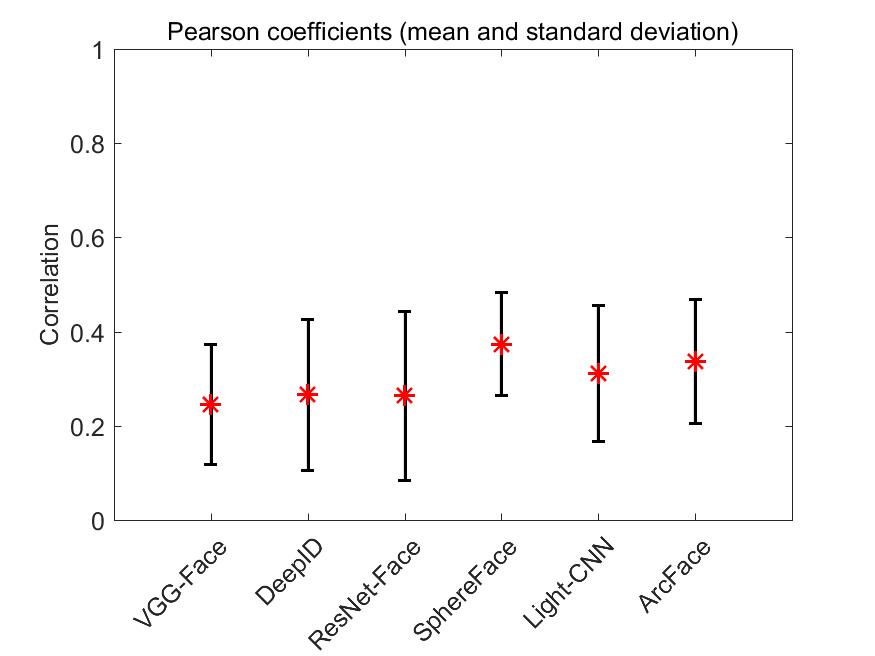}  }
\caption{
Pearson and Spearman coefficients between the outputted single neuron responses from all the referred CNNs and those by the linear model: (a) Pearson coefficients on LFW; (b) Spearman coefficients  on LFW; (c) Pearson coefficients on MegaFace; (d) Spearman coefficients on MegaFace.}   \label{corrLFWMEGAaxis}
\end{figure}

The mean values (also the standard deviations) of the Pearson and Spearman coefficients between the predicted representations by the axis model and those by all the referred CNNs are shown in Figure \ref{corrLFWMEGAaxis}. The significance of the computed correlations is also tested, and more than $80\%$ of the corresponding $p$-values  for each CNN are lower than the significance level of $0.01$. 
As seen from Figure \ref{corrLFWMEGAaxis}, both the computed coefficients for the six CNNs on the two datasets are close to  $0.25$ in most cases, in agreement with the above results for VGG-Face1/VGG-Face2 on Multi-PIE. This further suggests that the predicted representations  by the axis model are not strongly correlated with those of CNNs.

Figure \ref{reconMega} shows the reconstructed results on an examplar image from MegaFace,
which are also dramatically different from the original face image.

The $ACC$ and $EER$ on the two datasets by the axis model are shown in the green bars of Figure \ref{accLFWMEGA}.
All the predicted representations by the axis model give lower $ACC$ and $EER$ than the corresponding CNN representations. The verification results are similar to the  face recognition results on the third Multi-PIE dataset(as defined in Table \ref{fnet}), mainly because in these experiments, their training sets contains similar (even the same) poses to those in the testing sets, although the images in LFW and MegaFace have a mount of varying poses.

From all these results, we can see that the axis model is not as good as the linear model 
for modelling the neuron responses of DNNs.

\subsection{DNN neurons versus IT Neurons on face representation}

In \cite{Chang2017}, the following points on face representation in primate IT cortex are observed:
\begin{itemize}
\item[1] By formulating faces as points in a $50$-D parameter space,  human faces  could  be linearly decoded  from IT neuron responses,
and  the responses of IT neurons could be linearly predicted with the face vectors.
\item[2] The response of each face cell is the dot product of an incoming face vector onto its STA  axis, followed by a nonlinear rectification, called ``the axis model''. This model could  adequately decode face vectors from neural population responses and predict neural firing rates to new faces.

\end{itemize}

Comparing with the observations in IT cortex, the following points on CNNs are observed:
\begin{itemize}
\item[1] By formulating face images as points in a $50$-D parameter space, 
the face vectors could  also be linearly decoded from the representations at higher CNN layers, and  the representations at higher CNN layers could be linearly predicted with the face vectors. 
This indicates to a large degree, or  at a ``coarse-grained'' level,
CNNs have a similar linear encoding and decoding mechanism as that in primate IT cortex. 
\item[2]
The axis model fails to adequately model the face representations at higher CNN layers.
This suggests that  
the face representation mechanism in CNNs 
have noticeable discrepancies
with that in primate IT cortex at a ``finer-grained'' level,
similarly demonstrated for general object representations in \cite{Rishi2018}.
\end{itemize}

\section{Conclusions and discussions} \label{Consec}

In this work, we investigate the face representations of CNNs using six state-of-the-art CNNs as our model CNNs on three representative datasets, and our main findings are as follows:
\begin{itemize}
\item[\textcircled{1}]   CNNs for face recognition could be considered as a linear model in a $50$-D parameter space. Although the face representations of higher CNN layers are obtained  by implementing a cascade of nonlinear operators, these representations could in fact be encoded/decoded by the linear model in this parameter space, similar to primate IT cortex. 
Since all the six DNNs exhibit this linear encoding/decoding property and the six CNNs have diverse architectures, such as VGG-Face vs ResNet-Face, we thought this linear encoding/decoding property could not be due to some specific CNN architecture, but it should be an inherent 
property of face recognition DNNs in general.

\item[\textcircled{2}]  The linear model is more effective for modelling the face representations of CNNs than the axis model in \cite{Chang2017}, probably because the number of the fitted parameters  in the linear model is much larger than that  in the axis model. 

\item[\textcircled{3}] 
The face recognition and verification accuracies of the linear classifiers with the linearly-predicted representations as inputs are close to or even higher than those of the model CNNs. 
\end{itemize}

The above linear encoding  of face representation by CNNs in a parameter space seems both interesting and surprising, considering the parameter space is purely a mathematical concept and modern CNNs for face recognition, composed of many layers with enormous parameters to train, are in fact to recover a few dozen shape and appearance model parameters. What could be the implications of such a linear encoding for both deep learning and neurosciences? Here are some points:

\begin{itemize}
\item[\textcircled{1}] \textbf{The inverse generative model of CNNs:}
Currently, CNNs are largely of ``blackbox'' nature in the sense that their exceptionally good object recognition performances still lack sufficient explanatory theory. One of the proposals is called the inverse generative model \cite{Lin2017Why,Kulkarni2015Picture,Patel2015A}, that is, CNNs  are mainly to recover hierarchically the generative model parameters. The inverse graphics in \cite{NIPS2015_5851} and the hypothesis-and-verification approach in \cite{Yildirim2015Efficient}, are just such examples. The linear encoding in this work seems to support the theory of the inverse generative model, at least for face recognition. As linear encoding has quite a number of salient advantages as shown in \cite{Chang2017}, it seems worthy exploring new simpler networks to directly regress generative model parameters, which is also one of our future research directions, rather than to train a very-deep layered network by a heavy data-driven approach currently. 
Of course, how to establish an adequate parameterized model for a given class of objects is still a difficult research direction in both computer vision and computer graphics communities. 
\item[\textcircled{2}] \textbf{The goal-driven approach for sensory cortex understanding:}
Face recognition by CNNs, in essence, is purely data-driven under some recognition performance criteria. As shown in this work, CNNs have similar  linear face encoding mechanism to that by macaques. This seems to suggest that, the macaque face processing system could be modelled by only optimizing the face recognition performances of CNNs, which is in support to the goal-driven paradigm for sensory cortex understanding advocated in \cite{Yamins2014Performance,Yamins2016}.
\item[\textcircled{3}] \textbf{Validity of linear encoding  for familiar faces and faces with expressions:} 
It is generally believed that face recognition and face expression in primate are processed in different cortical areas, face recognition in IT, and face expression in the superior temporal
sulcus (STS) \cite{Rolls2017}. In \cite{Chang2017}, their axis model is mainly for rapid face recognition, or core face recognition \cite{Tsao2008Patches}. In addition, as shown in \cite{Landi2017Two}, two additional cortical areas are detected for only familiar face recognition in monkeys. Our results show that CNNs are able to handle both familiar and unfamiliar face images, as well as faces with different expressions. This seems to suggest that either monkey also has a linear encoding mechanism for familiar faces and faces with expressions, which needs to be clarified in the future, or the face encoding by CNNs has substantial differences with that in primate.  
\end{itemize}

Of course, our work also has some limitations notably:
\begin{itemize}
\item[\textcircled{1}]  This work only focuses on the face representation of CNNs, rather than general object representation of CNNs. 
Considering that different faces generally vary slightly in topology and geometry, while general objects (such as tables, chairs, cars, etc.) have no resemblance among them, whether this simple linear model for face representation is  extendable to general object modelling is doubtful. Besides, how to parameterize general objects seems also an insurmountable difficulty.

\item[\textcircled{2}]  There are various approaches for generating parameterized face images, other than our used AAM approach here, which could form different parameter spaces. Our results only reveal that there exists at least such a parameter space (determined by the AAM approach) where the face representations of CNN layers could be predicted by linearly encoding the face vectors.  In the future, other parameter spaces would be explored.
\item[\textcircled{3}]  In \cite{Szegedy2014}, it is reported that a distinct difference on object recognition between CNNs and human visual system is their sensitivity to adversarial images, that is, images slightly corrupted with random noise. Human visual system is generally immune to adversarial images,  while the performance of CNNs on object recognition is quite sensitive to them. It remains unclear whether the linear face representation mechanism in CNNs still holds on adversarial face images, which would be another line of our future works.
\end{itemize}

In summary, to the best of our knowledge, this work is the first attempt to  partially reveal the linear face representation mechanism in CNNs, different from commonly believed complex feature encoding by CNNs. In addition, our results shed some lights on the similarities and differences of face representation between CNNs and primate IT cortex.
Finally, our results reveal that the linear face encoding by CNNs might be used for designing new CNNs for face recognition, which is also one of our future research directions.

\section*{Data availability}
The CMU Multi-PIE dataset could be accessed at \url{http://www.cs.cmu.edu/afs/cs/project/PIE/MultiPie/Multi-Pie/Home.html}.
The LFW dataset could be accessed at \url{http://vis-www.cs.umass.edu/lfw/index.html}.
The MegaFace dataset could be accessed at \url{http://megaface.cs.washington.edu/}.

%\section*{Acknowledgements}
%This work was supported by the  National Natural Science
%Foundation of China (61573359).

%% use section* for acknowledgement
%\section*{Acknowledgment}

\bibliographystyle{splncs}
\bibliography{egbib}

\end{document}